%% file: acl_latex.tex
\pdfoutput=1
\documentclass[11pt]{article}

\usepackage[final]{acl}
\usepackage{times}
\usepackage{latexsym}

\usepackage[T1]{fontenc}
\usepackage[utf8]{inputenc}

\usepackage{microtype}

\usepackage{inconsolata}

\usepackage{graphicx}

\usepackage{booktabs}
\usepackage{multirow}
\usepackage{xcolor}
\usepackage{caption}
\usepackage{subcaption}
\usepackage{pgfplots}
\pgfplotsset{compat=1.18}
\usepackage{algorithm}
\usepackage{algpseudocode}
\usepackage{enumitem}
\usepackage{amsmath}
\usepackage{placeins}
\usepackage{float}

\newcommand{\shortnamep}{UniCoM}
\newcommand{\shortnamed}{CS-FLEURS}
\newcommand{\shortnamea}{SWORDS}

\title{UniCoM: A Universal Code-Switching Speech Generator}

\author{
 \textbf{Sangmin Lee\textsuperscript{1}},
 \textbf{Woojin Chung\textsuperscript{1}},
 \textbf{Seyun Um\textsuperscript{1}},
 \textbf{Hong-Goo Kang\textsuperscript{1}}
\\
\\
 \textsuperscript{1}Dept. of Electrical \& Electronic Engineering, Yonsei University, South Korea, Seoul
\\
 \small{
   \textbf{Correspondence:} \href{mailto:sangmin_lee@dsp.yonsei.ac.kr, woojinchung@dsp.yonsei.ac.kr, syum@dsp.yonsei.ac.kr, hgkang@yonsei.ac.kr}{\{sangmin\_lee, woojinchung, syum\}@dsp.yonsei.ac.kr, hgkang@yonsei.ac.kr}
 }
}

\begin{document}
\maketitle
\begin{abstract}

Code-switching (CS), the alternation between two or more languages within a single speaker's utterances, is common in real-world conversations and poses significant challenges for multilingual speech technology.
However, systems capable of handling this phenomenon remain underexplored, primarily due to the scarcity of suitable datasets.
To resolve this issue, we propose \textbf{Uni}versal \textbf{Co}de-\textbf{M}ixer (\shortnamep), a novel pipeline for generating high-quality, natural CS samples without altering sentence semantics. 
Our approach utilizes an algorithm we call \textbf{S}ubstituting \textbf{WORD}s with \textbf{S}ynonyms (\shortnamea), which generates CS speech by replacing selected words with their translations while considering their parts of speech.
Using \shortnamep, we construct \textbf{C}ode-\textbf{S}witching \textbf{FLEURS} (\shortnamed), a multilingual CS corpus designed for automatic speech recognition (ASR) and speech-to-text translation (S2TT). 
Experimental results show that \shortnamed~achieves high intelligibility and naturalness, performing comparably to existing datasets on both objective and subjective metrics.
We expect our approach to advance CS speech technology and enable more inclusive multilingual systems.
\end{abstract}

\section{Introduction}
\input{figure/fig_pipeline}
Multilingual speech technology has advanced rapidly in recent years. 
Early research primarily focused on similar language families~\cite{indicml} or a small range of languages~\cite{callhomeml}, but the latest developments over the recent few years have significantly increased model sophistication. 
Researchers have scaled models~\cite{whisper, usm} to support over 100 languages directly, or more than 1,000 languages~\cite{mms} through language-specific modules~\cite{adapter}.
However, most approaches are designed for monolingual utterances and struggle to effectively process speech containing multiple mixed languages.

In particular, code-switching (CS), the practice of alternating between languages within a conversation, is prevalent in multilingual communities.
However, research on code-switching speech technology, including automatic speech recognition (CS-ASR), has been limited, with few studies focused on performance improvement~\cite{cs2019, cs2024_1, cs2024_2}.
Moreover, code-switching speech-to-text translation (CS-S2TT) is even more underexplored, with minimal research addressing this task~\cite{css2ttapple, costa}.

Given this landscape, we identify two key questions: \textit{(1) Why has CS research lagged behind multilingual speech technologies?} and \textit{(2) Why are existing studies confined to a few major languages?}
We attribute these limitations primarily to the scarcity and accessibility of CS datasets. 
To elaborate, CS occurs at both the inter-sentential (between sentences) and intra-sentential (within a sentence) levels~\cite{csbook}, and constructing datasets for these cases, particularly intra-sentential CS with diverse language pairs, requires a substantial number of fluent multilingual speakers. 
However, their rarity and the high cost of data collection pose significant challenges.

In this context, there are two potential data augmentation methods to address these challenges: CS speech generation using a multilingual text-to-speech (TTS) system~\cite{gen_tts1, gen_tts2} or combining existing monolingual sources~\cite{gen_mix1, gen_mix2}.
The first approach generates natural CS samples by preserving speaker identity across utterances. However, scaling this method is challenging due to the complexities of massively multilingual TTS, such as resource imbalance and prosodic variation. 
The second approach synthesizes CS samples from abundant monolingual data but may lose linguistic nuances and speaker identity when utterances from different languages are simply concatenated, degrading the quality of intra-sentential CS.

To address these limitations, we propose \textbf{Uni}versal \textbf{Co}de-\textbf{M}ixer (\shortnamep), capable of generating intra-sentential CS samples across diverse languages while maintaining both contextual relevance and speaker consistency.
\shortnamep~consists of three stages: preprocessing, source mixing, and style unification.
In the preprocessing stage, potential artifacts are removed from the input sources to ensure the stability of the pipeline.
The source mixing stage introduces \textbf{S}ubstituting \textbf{WORD}s with \textbf{S}ynonyms (\shortnamea), a novel algorithm to generate speech-text pairs for intra-sentential CS.
Finally, the style unification stage standardizes speaker styles across different sources.
Consequently, building on~\shortnamep, we introduce \shortnamed: a comprehensive multilingual CS corpus encompassing 253 language pairs—marking a significant advancement over traditional single language pair datasets. 
Furthermore, the inclusion of over 70 $n$-way parallel sentences offers potential for future exploration in CS-S2TT tasks.
Experimental results demonstrate that~\shortnamed~achieves comparable intelligibility and naturalness to human-generated CS datasets.
We anticipate our approach will serve as a foundation for generalizable CS speech technology, paving the way for broader applications. 
Our contributions are summarized as follows:
\begin{itemize}[left=0pt]
    \item We introduce \shortnamep\footnote{\url{https://github.com/sanghyang00/unicom}}, a novel code-switching speech dataset generation pipeline that is applicable to a wide variety of languages.
    \item We propose \shortnamea, an algorithm for intra-sentential CS generation across diverse languages while preserving linguistic and speaker identity.
    \item We release \shortnamed, a first-ever, large-scale, and massively multilingual CS speech corpus.
\end{itemize}

\section{Related Work}
\subsection{Code-Switching Speech Dataset}
There are a few speech corpora that can be utilized for CS research. 
The Fisher corpus~\cite{fisher} and Spoken Wikipedia Corpus~\cite{spokenwiki} were not originally intended for CS but contain some CS data, making them useful for CS tasks. However, its applicability to CS is limited due to data scarcity.
The ASCEND corpus~\cite{ascend} includes both inter- and intra-sentential CS speech, facilitating the modeling of diverse CS patterns. However, its language coverage was limited to Mandarin-English, constraining applicability to broader multilingual settings.
While other CS speech corpora~\cite{miami, seame, arabiccs, mucs} exist, they shared the challenges related to limited resources and language spans.

\input{table/sequence_pos}
\subsection{Code-Switching Speech Generation}
\citet{gen_tts1} and \citet{gen_tts2} proposed a code-switching speech synthesis approach using multilingual TTS. 
The method translates text, aligns similar words between languages, and generates speech via TTS. 
However, scaling to a large number of languages remains challenging for TTS, restricting its applicability to major languages.

\citet{gen_mix1} and \citet{gen_mix2} generate inter-sentential code-switching samples by concatenating independent utterances from different monolingual speech sources. However, this approach struggles to produce intra-sentential code-switching data due to challenges in preserving the unique characteristics of each language, such as phrase structure, making it less representative of real-world code-switching scenarios.
Additionally, the lack of speaker consistency in concatenated samples raises concerns about speech naturalness.

\section{\shortnamep}
\vspace{-2pt}
In this section, we propose \textbf{Uni}versal \textbf{Co}de-\textbf{M}ixer (\shortnamep), a novel code-switching speech data generation pipeline.
\shortnamep~is a universal and generalized framework designed for broad applicability across a wide range of languages.
Unlike previous methods that generate inter-sentential code-switching samples by simply concatenating sentences from two different languages, \shortnamep~enables intra-sentential code-switching speech generation while preserving the unique linguistic features of each language, such as phrase structure and word order.
Moreover, \shortnamep~preserves speaker identity by unifying the speaker's style, making the generated speech more reflective of naturally occurring code-switching in real-life scenarios.
In the following sections, we provide a detailed description of each step in the proposed framework.

\vspace{-2pt}
\subsection{Preprocessing}
Before applying the generation model, we implemented two preprocessing strategies to refine the baseline data.
First, we identified that some samples in the existing monolingual dataset contained buzzing or white-noise-like artifacts, which could disrupt the subsequent generation pipeline.
Consequently, we observed that these artifacts significantly degraded sample quality after the voice conversion (VC) stage, resulting in crackling sounds.
To address this issue, we applied bandpass filtering to remove potential artifacts while preserving the majority of speech components.
Specifically, we set cutoff frequencies below 80 Hz and above 7000 Hz to eliminate unwanted noise while preserving speech integrity.
The second challenge we addressed was amplitude inconsistency.
In most speech corpora, particularly multilingual ones, variations in recording environments lead to differences in volume, with some samples significantly quieter or louder than others.
Moreover, we also observed that these amplitude variations adversely affected VC performance. 
To mitigate this issue, we applied amplitude normalization to ensure that all input speech samples were scaled uniformly.

\vspace{-2pt}
\subsection{Intra-Sentential Source-Mixing}
Compared to inter-sentential CS, intra-sentential CS requires a significantly more complex process.
Specifically, in inter-sentential CS, contextual consistency is less critical as language transitions occur at sentence boundaries. 
In contrast, intra-sentential CS happens within a sentence, making it essential to maintain contextual coherence. 
Therefore, a more sophisticated algorithm was necessary---one that goes beyond simply concatenating segments from source speech in different languages.

\vspace{-2pt}
\subsubsection{Substitution Strategy Selection}
Given these challenges, the first aspect we considered was how to generate an intra-sentential CS utterance while preserving the original sentence's meaning and unique linguistic characteristics.
To achieve this, we determined that replacing parts of a sentence with semantically equivalent segments from other languages would be an effective approach.
From this perspective, we explored two primary rearrangement methods: (1) phrase-level and (2) word-level substitution.

\vspace{2pt}
\noindent\textbf{Phrase-Level Substitution.} We initially hypothesized that phrase-level substitution would better preserve the CS ratio. However, this approach introduced structural issues due to cross-linguistic variation in phrase syntax. As shown in Table~\ref{tab_sequence}, semantically equivalent substitution of phrase-level segments often disregards language-specific grammar, yielding unnatural outputs.

\vspace{2pt}
\noindent\textbf{Word-Level Substitution.} In contrast, part-of-speech (POS)—a word-level feature—is cross-linguistically common~\cite{pos_uni}, enabling substitutions that preserve both naturalness and meaning. We thus adopted word- or POS-level substitutions with a flexible number of substitutions, allowing natural CS sentence generation while preserving semantics and syntactic structure.

\vspace{-2pt}
\subsubsection{SWORDS Algorithm}
\label{sec:swords}
In this context, we propose the \textbf{S}ubstituting \textbf{WORD}s with \textbf{S}ynonym (\shortnamea) algorithm.
\shortnamea~is composed of four steps: (1) Sampling of equivalent sentence pairs, (2) Wordpair mapping generation, (3) Segmentation using forced alignment, (4) Completion through recombination.

This approach introduces two key properties.
First, \shortnamea~is the first method tailored for intra-sentential CS speech generation, whereas existing approaches were designed for inter-sentential CS.
Second, \shortnamea~preserves language-specific structure and sentence semantics through linguistically informed substitution, yielding natural outputs which highly resemble real-world CS scenarios.
Details of each step are provided in the following, and the pseudo-code is shown in Alg. \ref{alg:intra}.

\vspace{2pt}
\noindent\textbf{Sampling of Equivalent Sentence Pairs.}
We first organized $n$-way parallel sentences with equivalent meanings from a multilingual speech-text dataset. 
Next, we constructed one-to-one sentence-level mapping pairs for every language combination, promoting the word pair mapping generation process.

\vspace{2pt}
\noindent\textbf{Wordpair Mapping Generation.}
Next, we decomposed the sentence pairs with equivalent meanings into word pairs with corresponding meanings.
These word pairs were then classified based on the part of speech (e.g., noun, verb, adverb, adjective, and interjection).
This process was conducted using a large language model (LLM), GPT-4o-mini~\cite{gpt4} as the word-level mapping generator. 
By inputting the previously constructed sentence pairs into the LLM with input prompts, we generated a dictionary of equivalent word pairs organized by POS.
The details of the prompt and processing steps are illustrated in~\ref{sec:detail_llm}.

\input{table/dataset_size}
\vspace{2pt}
\noindent\textbf{Segmentation Using Forced Alignment.}
Consequently, to extract speech segments for each word pair generated from the previous step, we leveraged MMS-FA~\cite{mms} for forced alignment between text and speech, clipping each sample to match the target words.
Although MMS-FA requires Romanization, it offers an effective trade-off between alignment speed and accuracy compared to Whisper-based ones~\cite{whisper, whisperx}.
Details of the forced alignment methods we considered are provided in~\ref{sec:fa}.

\vspace{2pt}
\noindent\textbf{Completion Through Recombination.}
Finally, we rearranged the clipped segments to produce the final CS speech sample. 
During the rearrangement process, we randomly select the matrix language, with the other language acting as the embedded language. 
In the final generation phase, a predefined number of words are randomly selected from the word pairs and substituted into the matrix language utterance, resulting in a source-mixed sample.
Specifically, \shortnamep~enables the selection of various combinations of part-of-speech categories and word pairs for substitution, offering flexibility as a hyperparameter. 
This allows users to adjust the pool of part-of-speech categories and the number of substitutions based on specific requirements.

\input{pseudocode/intra}
\vspace{-2pt}
\subsection{Style Unification}
Since intra-sentential CS occurs within a sentence, preserving the speaker's identity throughout the utterance is essential. 
In the style unification stage, we aimed to align speaker styles across the segments, which were sampled from different utterances to ensure the naturalness of the generated samples. 
Particularly, our objective was to make the samples as close as possible to real-world instances of intra-sentential CS, with emphasis on enhancing speaker similarity.

Accordingly, we adopt kNN-VC~\cite{knnvc} as the voice conversion (VC) model for two main reasons.
First, kNN-VC retrieves ground-truth self-supervised features of the target speaker using a k-nearest neighbor (kNN) algorithm, uniquely distinguishing it from other VC models.
Given that the neural vocoder is relatively language-invariant, this retrieval-based mechanism preserves intelligibility while effectively transferring speaker characteristics across languages.
As a result, kNN-VC produces highly intelligible and natural speech, even in cross-lingual settings.
Since our VC pipeline targets cross-lingual scenarios, kNN-VC serves as an effective choice for unifying speaker identity across concatenated segments.

Second, kNN-VC achieves fast inference during the voice conversion process, primarily due to its adoption of the lightweight GAN-based vocoder~\cite{hifigan}, in combination with the retrieval-based method described earlier. This design enables substantially faster voice conversion compared to approaches that rely on diffusion-based methods or employ larger vocoders, while still maintaining competitive performance.
Although such alternatives may offer higher fidelity in certain scenarios, kNN-VC provides a more favorable trade-off between synthesis quality and inference speed. As a result, kNN-VC is well-suited for both online and offline generation scenarios required by \shortnamep. Further details behind the selection of the VC model are provided in Appendix~\ref{sec:vc_selection}.

\vspace{-2pt}
\section{CS-FLEURS}
\vspace{-2pt}
This section describes \shortnamed, an intra-sentential CS speech dataset spanning 253 language pairs, constructed using \shortnamep~with additional methods to enhance data quality.

\vspace{-2pt}
\subsection{Baseline Dataset}
To select the baseline dataset with a sufficient range of languages, we considered three datasets: CommonVoice~\cite{cv}, FLEURS~\cite{fleurs}, and FLEURS-R~\cite{fleursr}.

\noindent\textbf{CommonVoice} is a crowdsourced multilingual dataset covering over 120 languages and 30,000 hours. While its scale is a strength, variations in recording conditions often result in noisy samples.

\noindent\textbf{FLEURS} is a 102-language multilingual speech-text dataset with two key advantages over CommonVoice.
First, it supports both ASR and S2TT via $n$-way parallel sentences, enabling CS-ASR/S2TT dataset creation.
Second, its longer average sample length benefits the VC module, which relies on sufficient reference duration.

\noindent\textbf{FLEURS-R} is a denoised version of FLEURS generated using a speech restoration model~\cite{miipher}, retaining the original structure while improving clarity and reducing noise.
Since noise is a major obstacle in ensuring the quality of VC, we opted to use FLEURS-R as the baseline dataset.

\subsection{Language Selection}
\label{langselec}
We leveraged 23 languages from FLEURS-R (a subset focusing on European languages) that overlap with the VoxPopuli dataset's~\cite{voxpopuli} language coverage. We refer to these overlapping languages as in-domain (ID) languages.
The rationale behind this selection is that, although kNN-VC generates converted speech by retrieving ground-truth speaker embeddings, typological differences between source and target languages might lead to unnatural-sounding speech due to mismatched phonetic and prosodic characteristics across languages.
To mitigate such artifacts, we restricted the coverage of \shortnamed~to cover the languages that exhibit similar phonetic and orthographic properties (mostly Indo-European or Latin character languages) as discussed in previous linguistic studies~\cite{langsim1, langsim2}, to ensure more consistent, intelligible, and natural-sounding voice conversion results.
While samples from out-of-domain (OOD) languages often produced plausible outputs, we conservatively limited our evaluation to in-domain (ID) languages to ensure overall quality and stability.

\vspace{-2pt}
\subsection{Source Mixing}
\vspace{-1pt}
In Sec.~\ref{sec:swords}, we demonstrated that \shortnamep~can substitute various POS and adjust the number of word pairs as needed. 
However, excessive word substitution can lead to unnatural results, deviating from real-world CS scenarios.
To maintain naturalness and ensure the generated speech reflects authentic CS patterns, we limited sampling to a maximum of three word pairs and restricted POS categories to nouns, verbs, and interjections, following prior linguistic studies on natural CS~\cite{csnat1, csnat2, csnat3}.

\input{table/csfleurs_example}
\vspace{-4pt}
\subsection{Dataset Statistic}
\vspace{-1pt}

% \vspace{2pt}
\noindent\textbf{Data Size and Language Span.}
Recent trends in deep learning models underscore the importance of scaling up dataset size. 
A larger and more balanced dataset contributes to better generalization performance, allowing models to learn richer and more nuanced representations. 
In this context, \shortnamed~presents a significant advantage. 
It encompasses over 250 diverse language pairs and features an extensive amount of speech data. 
Although individual language pairs contain modest amounts of data, the broad linguistic coverage and overall scale make \shortnamed~a valuable resource for CS research in massively multilingual settings.

\vspace{2pt}
\noindent\textbf{Code-Switching Metrics.}
The code-mixing index~\cite{cmi} (CMI) is a widely used metric for quantifying the intensity and proportion of code-mixing within a given text.
It measures the ratio between the matrix (primary) language and the embedded (inserted) language. A higher CMI indicates a greater extent of CS.
The I-index~\cite{i-index}, on the other hand, indicates the ratio of the occurrence points of CS within the utterance or text.
It aims to measure how evenly CS occurs throughout the speech sample.
Unlike previous CS speech corpora that did not provide such information, \shortnamed~presents the distributions of both CMI and I-index, as shown in Figure~\ref{fig:cmi} and \ref{fig:i_index}.
Moreover, language types (e.g., matrix, embedded) and metrics are incorporated into the metadata of \shortnamed, enabling it to be a valuable tool for uncovering the linguistic characteristics of CS and exploring its relationship with speech patterns in future research. To support qualitative analysis, Table~\ref{cs_fleurs_example} shows an annotated transcription pair from \shortnamed.

\vspace{2pt}
\noindent\textbf{Phoneme Diversity.}
Quantifying phoneme diversity in a multilingual corpus is inherently challenging, and this task becomes even more complex when applied to \shortnamed, a code-switching corpus. 
To address this, we applied Romanization~\cite{uroman} uniformly across all textual data based on prior studies showing that Romanization preserves phonetic characteristics to a reasonable extent~\cite{mms, phontoken}. 
By leveraging Romanized character diversity, we approximate phoneme diversity, even if not perfectly.
When compared to the baseline dataset, \shortnamed~functions as a code-switching speech corpus without compromising phoneme diversity, as shown in Figure~\ref{fig:phoneme_diversity}.
This feature is particularly useful for training speech-to-text models, where phoneme diversity is crucial for performance.

\input{figure/distribution}
\vspace{-1pt}
\section{Experiments}
\vspace{-2pt}
\input{table/cs_fleurs_full}
In this section, we evaluate the validity of CS-FLEURS to prove the effectiveness of \shortnamep.
All experiments were performed on two RTX 3090 GPUs with 24GB VRAM each, and subjective metrics were evaluated through a user study involving a total of 42 participants and over 2,500 utterances.

\vspace{-6pt}
\subsection{Evaluation Metrics}
\vspace{-2pt}
\noindent\textbf{Romanized Character Error Rate (RER).}
Given the extensive multilingual capabilities and CS nature of \shortnamed, evaluating a unified metric for all combinations was challenging due to the lack of trainable open-source models for massively multilingual CS-ASR.
To resolve this issue, we first converted all transcriptions into Romanization pairs and measured the Romanized character Error Rate (RER) to assess intelligibility, based on the previous studies that proved Romanization standardizes orthographic diversity while preserving phonetic characteristics~\cite{mms, phontoken}. This characteristic makes RER analogous to measuring Phoneme Error Rate (PER), where a lower RER indicates high intelligibility.
Specifically, we fine-tuned the XLS-R~\cite{XLS-R} using the CommonVoice 17.0 dataset for our Romanization-based ASR model.

\vspace{2pt} 
\noindent\textbf{Mean Opinion Score (MOS).}
Since \shortnamep~generates CS speech through VC, ensuring the naturalness of the generated samples is crucial.
To this end, we adopted MOS, a subjective metric to evaluate perceptual speech quality where listeners rate the speech on a scale from 1 to 5, with higher scores indicating more natural and intelligible speech.

\vspace{2pt} 
\noindent\textbf{Speaker Identity Score (SIS).}
Given the intra-sentential CS nature, it is essential to ensure that speaker identity is preserved within each sample.
To evaluate this aspect, we introduce the Speaker Identity Score (SIS), a metric designed to assess the consistency of speaker identity. 
The SIS is measured similarly to MOS, where raters evaluate the confidence that a given sample contains only a single speaker. 
The score also ranges from 1 to 5, reflecting the perceived preservation of speaker identity within the sample.

\vspace{-3pt}
\subsection{Dataset Quality Evaluation}
\noindent\textbf{Intelligibility Evaluation.} In Table~\ref{tab_cs_fleurs_full}, results demonstrate that CS-FLEURS maintains sufficient intelligibility, with an average RER of 31\%.
Given that CS speech is less intelligible than monolingual speech and \shortnamed~originates from a VC pipeline, results highlight the potential of \shortnamep~to mitigate CS speech corpus scarcity.

\vspace{2pt}
\noindent\textbf{Naturalness Evaluation.} Table~\ref{tab_cs_fleurs_full} also indicates that samples from \shortnamed~show only minor deviations from human-generated datasets in terms of human perception. 
Specifically, MOS values generally exceed 4, even for the lowest-scoring languages, while samples from \shortnamed~maintain high speaker consistency, with SIS scores exceeding 4.5 for the majority of language pairs.

\vspace{-3pt}
\subsection{Comparative Evaluation}
\input{table/full_eval_merged}
Moreover, to validate the practicality of \shortnamed, we conduct comparisons with existing human-annotated CS datasets across both ID and OOD languages. For ID languages, we used the Spoken Wikipedia Corpus (SWC) and the Miami-Bangor Corpus (MBC), while for OOD languages, we utilized the ASCEND and MUCS2021 corpora. We considered only CS samples from each dataset and evaluated their relative quality using the same metrics described in previous sections.
Especially, for OOD languages, the quality of generated datasets might be inevitably degraded due to the large linguistic difference as mentioned in Section~\ref{langselec}. To distinct it from the original \shortnamed~and ensure a fairer comparison, we categorize them separately as \shortnamed-O.

\vspace{2pt} 
\noindent\textbf{In-Domain Languages.} As illustrated in Table~\ref{tab_id}, \shortnamed~exhibits competitive intelligibility compared to human-generated CS datasets in ID languages. Notably, \shortnamed~achieved comparable RER and MOS scores to SWC, a professionally produced and well-aligned dataset. Moreover, \shortnamed~outperformed MBC, which is a noisy dataset, in both objective and subjective evaluations. These findings highlight the quality of \shortnamed~and prove the strength of \shortnamep.

\vspace{2pt}
\noindent\textbf{Out-of-Domain Languages.} Table~\ref{tab_ood} shows that \shortnamed~achieves performance comparable to existing datasets under OOD conditions, with slightly lower MOS but showing better RER scores. For SIS, it performs on par with human-generated CS corpora. These results underscore the generalizability of \shortnamep, despite the typical degradation observed in OOD compared to ID settings.

\vspace{-2pt}
\subsection{Impact of \shortnamed~in CS-ASR}
Finally, to evaluate the actual contribution of \shortnamed~to CS-ASR performance, we conducted experiments using the aforementioned two ID human-generated CS datasets along with corresponding language pairs in CS-FLEURS (CSF). We ensured an equal-sized training set for each experiment and fine-tuned XLS-R with a concatenated vocabulary for our CS-ASR model.

\vspace{2pt}
\noindent\textbf{English-German Pair.} As shown in Table~\ref{tab_cs_impact}, training the CS-ASR model with \shortnamed~improves transcription performance on the English-German pair. In particular, evaluation on the SWC dataset shows that combining our synthetic data yields better results than using SWC alone, highlighting its value as a data augmentation method. Moreover, when evaluated on its own, \shortnamed~also proves effective as a primary training resource.

\vspace{2pt}
\noindent\textbf{English-Spanish Pair.} 
Similarly, when training solely on MBC for the English-Spanish pair, the high noise level results in models producing only blank tokens.
In contrast, joint training with \shortnamed~substantially improves performance on MBC, demonstrating its effectiveness for data augmentation. Notably, training exclusively on the synthetic data yields the best results on the corresponding evaluation set (CSF), reinforcing its capacity both as a core training source and as a means of enhancing existing datasets.
However, both tables reveal performance degradation in cross-dataset inference due to domain mismatch across datasets, as previously discussed by~\citet{whisper}. 
To summarize, results for both language pairs highlight the strength of \shortnamed~in two key aspects: (1) as a strong standalone training dataset for CS-ASR, and (2) as an effective data augmentation method for existing CS-ASR corpora.

\input{table/impact_csasr}
\vspace{-2pt}
\section{Conclusion}
\vspace{-2pt}
In this paper, we introduce \shortnamep, a universal pipeline for code-switching speech generation.
Specifically, we propose a \shortnamea~algorithm to facilitate the generation of intra-sentential code-switching samples while maintaining the natural meaning of the original speech.
Our pipeline then employs voice conversion to ensure the naturalness of generated samples by unifying speaker styles across utterances without compromising intelligibility.
Finally, we release \shortnamed, a massively multilingual code-switching speech dataset designed for the CS-ASR task while offering future utility for CS-S2TT scenarios.
In experiments, \shortnamed~exhibits high intelligibility and naturalness compared to human-generated datasets, while contributing to the improvements in CS-ASR training and performance.
We believe our work will serve as a cornerstone for a more generalized code-switching speech technology in the future.

\clearpage
\section{Limitations}
Since the generation pipeline includes a pre-trained voice conversion model, the performance of \shortnamep~is inevitably limited for out-of-domain languages for the voice conversion model. 
Consequently, to generate a high-quality and natural CS dataset, the language span of \shortnamed~is constrained to combinations of languages within the in-domain set of the voice conversion model.
Notably, in terms of orthographic features, \shortnamed~includes languages that predominantly use the Latin script. 
While Bulgarian leverages the Cyrillic script and Greek utilizes the Greek alphabet, the remaining languages are based on the Latin script.

In future research, we aim to expand the scope of \shortnamep~by constructing a multilingual voice conversion model, enabling the application of \shortnamep~to languages with diverse orthographies.

\section{Ethical Statements}
This study follows ethical guidelines, prioritizing fairness, transparency, and accountability.

The code for \shortnamep~and demo version of \shortnamed~is publicly available on anonymous GitHub via the footnote link, and the datasets used for validating \shortnamep~and \shortnamed~(e.g., CommonVoice, FLEURS-R, ASCEND, MUCS2021) are fully open-source.
\shortnamed~follows the Creative Commons Attribution 4.0 (cc-by-4.0) license, in accordance with FLEURS-R, and will be publicly released after undergoing post-processing of metadata.

We ensured consistency and fairness in comparisons across all experiments. 
In cases where fair comparisons were not possible due to specific conditions, we added the $\dag$ token to indicate this to potential readers.
For the user study, all participants voluntarily participated through community outreach. 
Each participant was thoroughly informed about the study process, procedures, and the intended use of the results. 
The payment was appropriate given the participants' demographics.

We recognize the impact of massively multilingual code-switching speech research and are committed to conducting our work while adhering to ethical research practices.

% Bibliography entries for the entire Anthology, followed by custom entries
%\bibliography{anthology,custom}
% Custom bibliography entries only
\clearpage
\bibliography{custom}

\clearpage
\appendix
\section*{\LARGE{Appendix}}
\section{Technical Details of SWORDS}
\vspace{-2pt}
In this section, we present additional details of the \shortnamea~algorithm, covering the generation of word-level mappings using LLMs, along with the input prompts and post-processing steps, and the selection criteria in the forced alignment module.

\vspace{-2pt}
\subsection{Word-Level Mapping Generation}
\vspace{-2pt}
\label{sec:detail_llm}
% \vspace{4pt}
\noindent\textbf{Hyperparameter of LLM.}
For hyperparameters of LLM, we set the temperature to 0.0 to produce fully deterministic results and to ensure output validity in word-level mapping.
All other hyperparameters were kept at their default values.

\vspace{4pt}
\noindent\textbf{Input Prompt.}
For the input prompt, we utilized two system prompts and one user prompt, as shown in Tab~\ref{prompts}. 
The system prompts included a role-playing prompt to enhance language-specific performance and a formatting prompt to ensure a consistent output structure. 
The user prompt was designed to generate word-level mappings based on the input language and text.
However, due to the complexity of the desired output—a YAML-formatted dictionary with nested lists—occasional inconsistencies in the output format were observed.

\vspace{4pt}
\noindent\textbf{Post-Processing.}
To address this issue, we applied post-processing to enforce a consistent data structure across all LLM outputs. 
Specifically, we removed unpaired elements, intentionally inserted missing keys, and converted different data structures (e.g., dictionaries, tuples) into lists.
This post-processing ensured that all LLM outputs maintained a consistent and valid word-level mapping structure, which was then used in the forced alignment process.
The overall word-level mapping generation process is illustrated in Figure~\ref{fig:wordmap}.

\vspace{-2pt}
\subsection{Selection of Forced Alignment Model}
\label{sec:fa}
\noindent\textbf{Whisper-Based Models.}
The first approach we considered was utilizing methods based on Whisper, a multilingual ASR model.
To achieve a superior alignment performance, we used the largest Whisper model, which was crucial for maintaining high-quality output in our final dataset. 
However, as the model size increased, the alignment time grew substantially, making real-world applications difficult.
Although we considered using a smaller model, this resulted in a decline in alignment performance, which was critical for us, as the quality of the generated results was our top priority.  

\noindent\textbf{MMS-FA.}
As an alternative, we employed MMS-FA with default hyperparameters (e.g., clipping threshold), which offers forced alignment with a smaller model size and full GPU-based computation. Although it requires orthographic unification, MMS-FA provides faster and effective alignment compared to the Whisper-based methods, making it more suitable for real-world applications.

\vspace{-3pt}
\section{Language-Pair-Specific Result}
\vspace{-3pt}
\label{sec:lang_spec}
In this section, we present a detailed analysis of the experimental results for each language combination, providing insights into the system's performance across different language pairs. 
The full results for 253 language pairs are shown in Table~\ref{appendix_full_pt1} and Table~\ref{appendix_full_pt2}, while the distributions of metrics across the language pairs are illustrated in Figure~\ref{fig:all_distributions}.
These analyses would help evaluate performance variations and identify emerging patterns.

\vspace{-3pt}
\section{Selection of Voice Conversion Model}
\vspace{-3pt}
\label{sec:vc_selection}
In this section, we elaborate on the criteria used to select the VC module for \shortnamep. Given our objective, we considered two key factors: (1) sufficient intelligibility in cross-lingual settings and (2) fast generation speed. We placed greater emphasis on the latter, as slower generation can substantially increase the overall construction time, especially for large-scale datasets. To assess the suitability of each model for \shortnamep, we conducted an experiment comparing different voice conversion models. Specifically, we evaluated two additional VC models—Diff-HierVC~\cite{diffhiervc} and SeedVC~\cite{seedvc}—in cross-lingual voice conversion scenarios. Both adopt decomposition-based (e.g., pitch, content, etc) approaches, in contrast to the retrieval-based architecture of kNN-VC.

Results in Table~\ref{tab_abl_vc} show that kNN-VC satisfies both criteria effectively. While it exhibited slightly lower intelligibility and naturalness compared to the diffusion-based alternatives, the real-time factor (RTF)—which measures the time required to process one second of audio—demonstrates that diffusion-based models are significantly slower than kNN-VC, regardless of whether a lightweight~\cite{hifigan} or heavyweight~\cite{bigvgan} vocoder is used.
We considered such minor degradations acceptable, as utterances in the embedded language of code-switched speech often exhibit imperfect pronunciation or reduced intelligibility by nature. Based on this trade-off, we chose to adopt kNN-VC.
\input{table/appendix_prompt}
\input{figure/fig_wordmap}
\input{table/appendix_full}

\input{table/appendix_full2}

\input{figure/appendix_distribution}
\input{table/abl_vc}
\end{document}

%% file: figure/fig_pipeline.tex
\begin{figure*}[!t] 
    \centering
    \includegraphics[width=\textwidth]{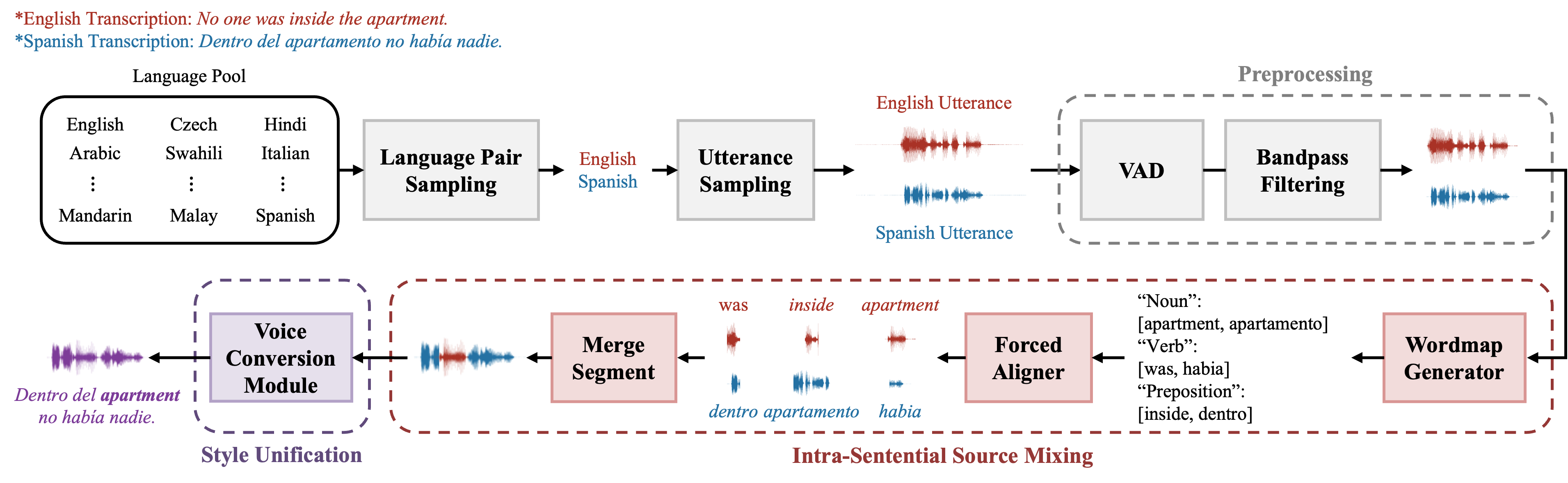}
    \caption{The overall pipeline of UniCoM. The bold region of the final transcription denotes the code-switching.}
    \label{fig:pipeline} 
    \vspace{-5pt} 
\end{figure*}

%% file: table/sequence_pos.tex
\begin{table*}[]
\centering
\small
\begin{tabular}{c|ccl} \toprule
Order & Example        & Frequency (\%) & Languages                                                                                                                                                                                                                                                                                                                                                                                                                                                                                \\ \midrule\midrule
SOV   & Sam \textcolor{blue}{apples} \textcolor{red}{ate} & 45             & \begin{tabular}[c]{@{}l@{}}Abaza, Abkhaz, Amharic, Akkadian, Armenian, Azerbaijani, Basque, etc.\end{tabular} \\ \midrule
SVO   & Sam \textcolor{red}{ate} \textcolor{blue}{apples} & 42             & \begin{tabular}[c]{@{}l@{}}Chinese, many European languages, Swahili, Thai, Vietnamese, etc.\end{tabular}                                                                                                                                                                                                                                                                                                     \\ \midrule
VSO   & \textcolor{red}{Ate} Sam \textcolor{blue}{apples} & 9              & \begin{tabular}[c]{@{}l@{}}modern Arabic, Berber languages, Filipino, Irish, Māori, Welsh, etc.\end{tabular}                                                                                                                                                                                                                                                                                                                     \\ \midrule
VOS   & \textcolor{red}{Ate} \textcolor{blue}{apples} Sam & 3              & \begin{tabular}[c]{@{}l@{}}Austronesian languages, Car, Chumash, Fijian, Malagasy, etc.\end{tabular}                                                                                                                                                                                                                                                              \\ \midrule
OVS   & \textcolor{blue}{Apples} \textcolor{red}{ate} Sam & 1              & Äiwoo, Hixkaryana, Urarina                                                                                                                                                                                                                                                                                                                                                                                                                \\ \midrule
OSV   & \textcolor{blue}{Apples} Sam \textcolor{red}{ate} & 1 $\downarrow$   & Tobati, Warao, Haida  \\ \bottomrule                                                                                                                                                                                                                                                                                                                                                                                                                                                              
\end{tabular}
\caption{A comparison of phrase order across languages~\cite{order2, order1}. Mixing sources with "ate apples" as the target may result in the loss of linguistic characteristics due to phrase order variations.}
\label{tab_sequence}
\vspace{-7pt} 
\end{table*}

%% file: table/dataset_size.tex
\begin{table*}[!t]
\small
\centering
\begin{tabular}{l|cccccccc} \toprule
Dataset   & \# language pair & duration (h)      & \# utt     & \# tok    & CMI & I-index & \# parallel \\ \midrule
Spoken Wikipedia\textsuperscript{$\dagger$}    &      1 (en-de) &       6.0       &      2.5k      &     10.9k      & - & - & \textit{N/A}\\
Miami-Bangor\textsuperscript{$\dagger$}    &      1 (en-es) &       5.0       &      2.3k      &     5.5k      & - & - & 2\\
ASCEND    & 1 (en-zh)     &      10.6        &      12.3k      &      11.4k     & - & - & \textit{N/A}\\
MUCS2021  &      2 (bn/hi-en) &       148.2       &      86.8k      &      27.9k     & - & - & \textit{N/A}\\ \midrule
\textbf{CS-FLEURS} &     \textbf{253} &       \textbf{1.6k}       &     \textbf{654.7k}       &      \textbf{179.6k}     & 0.11 & 0.19 & 73 \\ \bottomrule
\end{tabular}
\caption{Overall comparison between human-generated CS speech datasets and CS-FLEURS. CMI and I-index values are scaled between 0 and 1, $\dagger$ indicates that CS samples are partial, and statistics reflect only the CS subset.}
\label{tab_size}
\vspace{-7pt} 
\end{table*}

%% file: pseudocode/intra.tex
\setlength{\textfloatsep}{10pt}
\begin{algorithm}[!t]
\caption{SWORDS algorithm.}
\label{alg:intra}
\begin{algorithmic}[1]
\small\Statex \textbf{Input:} Multilingual speech-text paired dataset $\mathcal{D}$, Number of substitutions $N$, POS substitution categories $\mathcal{P}$
\small\Statex \textbf{Output:} $SWORDS(\mathcal{D}$, $N$, $\mathcal{P})$
% \small\Procedure{SWORDS}{$\mathcal{D}$, $N$, $\mathcal{P}$}
    \small\State \textcolor{gray}{// Step 1: Sample equivalent sentence pairs}
    \small\State $S_{pairs} \gets \text{SampleEquivalentSentences}(\mathcal{D})$
    \small\State $lang_1, lang_2 \gets \text{SelectLanguagePair}(S_{pairs})$
    \small\State $utt_{1}, txt_{1} \gets \text{SelectSample}(lang_1)$
    \small\State $utt_{2}, txt_{2} \gets \text{SelectSample}(lang_2)$
    \\
    \small\State \textcolor{gray}{// Step 2: Wordpair mapping generation}
    % \small\For{each $(s_1, s_2)$ in $S_{pairs}$}
    \small\State $W_{pairs} \gets \text{LLMWordMapping}(txt_1, txt_2)$
    \small\State $W_{pairs} \gets \text{ClassifyByPOS}(W_{pairs})$
    \small\State $W_{sub} \gets \text{RandomSelect}(W_{pairs}, N, \mathcal{P})$
    % \small\EndFor
    \\
    \small\State \textcolor{gray}{// Step 3: Forced alignment segmentation}
    \State $segment_{sub} \gets$ [ ]
    \small\For{each word $w$ in $W_{sub}$}
        \State $segment_w \gets \text{MMS-FA}(w, utt_1, utt_2)$
        \State Append $segment_w$ to $segment_{sub}$
        % \State $W_{segments} \gets W_{segments} \cup \{\text{segment}_w\}$
    \small\EndFor
    \\
    \small\State \textcolor{gray}{// Step 4: Recombination}
    \small\State $lang_{mat} \gets \text{RandomSelect($lang_1, lang_2$)}$
    \small\For{each word $w$ in $W_{sub}$}
        \State $txt_{cs} \gets \text{Substitute}(word, txt_1, txt_2, lang_{mat})$
    \small\EndFor
    \small\For{each $seg$ in $segment_{sub}$}
        \State $utt_{cs} \gets \text{Substitute}(seg, utt_1, utt_2, lang_{mat})$
    \small\EndFor
    % \small\State $utt_{cs}\gets \text{SubstituteSeg}(\text{$utt_1$, $utt_2$}, \text{$word_{sub}, lang_{mat}$})$
    % \small\State $\hfill \gets \text{SubstituteSeg}(\text{$utt_1$, $utt_2$}, \text{sub\_words})$
    % \small\State $txt_{cs} \gets \text{SubstituteSeg}(\text{$txt_1$, $txt_2$}, \text{$word_{sub}, lang_{mat}$})$
    % \small\State $\hfill \gets \text{SubstituteSeg}(\text{$txt_1$, $txt_2$}, \text{sub\_words})$
    \\
    \\
\small\Return $utt_{cs}, txt_{cs}$
% \small\EndProcedure
\end{algorithmic}
\end{algorithm}

%% file: table/csfleurs_example.tex
\begin{table}[!t]
\centering
\small
\resizebox{\columnwidth}{!}{%
\begin{tabular}{l|l} \toprule
            & Transcription \\ \midrule\midrule
English    & \begin{tabular}[l]{@{}l@{}}Hiking is an outdoor activity which consists of\\walking in natural environments, often on hiking trails.\end{tabular}           \\ \midrule
Dutch    & \begin{tabular}[l]{@{}l@{}}Wandelen is een buitenactiviteit waarbij je in een\\natuurlijke omgeving wandelt, meestal op wandelpaden.\end{tabular}           \\ \midrule
\begin{tabular}[l]{@{}l@{}}Code-\\Switched\end{tabular}   & \begin{tabular}[l]{@{}l@{}}\textcolor{red}{Wandelen} \textit{is an outdoor activity which consists of}\\\textit{walking in natural environments often on} \textcolor{red}{wandelpaden}.\end{tabular}         \\ 
\bottomrule
\end{tabular}
}
\caption{Example transcription pair from CS-FLEURS, with the matrix language in italics and code-switched (or embedded language) segments highlighted in red.}
\label{cs_fleurs_example}
\end{table}

%% file: figure/distribution.tex
\begin{figure}[t]
    \centering
    \begin{subfigure}[b]{0.5\textwidth}
        % \centering
        \begin{tikzpicture}
            \begin{axis}[
                width=\textwidth, 
                height=3.8cm, 
                ylabel={Density (\%)}, 
                ylabel style={font=\small, yshift=-7pt}, 
                xmin=0, xmax=0.5, 
                ymin=0, ymax=10, 
                xtick={0, 0.1, 0.2, 0.3, 0.4, 0.5},
                ytick={0, 5, 10},
                xtick pos=bottom,
                xticklabel style={font=\scriptsize, /pgf/number format/fixed}, 
                ytick pos=left,
                yticklabel style={font=\scriptsize, /pgf/number format/fixed}, 
                grid=major, 
                xmajorgrids=false,
                bar width=0.01,
                legend style={at={(0.98, 0.95)}, anchor=north east, font=\small},
            ]
            
            \addplot[ybar, fill=blue, opacity=0.7] coordinates {
                (0.01, 0.0254) (0.03, 1.9053) (0.05, 8.2082) (0.07, 8.3602) (0.09, 6.4173)
                (0.11, 6.2746) (0.13, 5.1549) (0.15000000000000002, 3.7387) (0.16999999999999998, 2.1617) (0.19, 1.9738)
                (0.21000000000000002, 1.8224) (0.22999999999999998, 0.9640) (0.25, 0.9057) (0.27, 0.4412) (0.29000000000000004, 0.3190)
                (0.31, 0.3543) (0.33, 0.3630) (0.35, 0.0121) (0.37, 0.0716) (0.39, 0.1446)
                (0.41000000000000003, 0.0936) (0.43, 0.0880) (0.45, 0.0469) (0.47, 0.0075) (0.49, 0.0002)
                (0.51, 0.0790) (0.53, 0.0009) (0.55, 0.0037) (0.5700000000000001, 0.0181) (0.59, 0.0008)
                (0.61, 0.0171) (0.63, 0.0037) (0.65, 0.0015) (0.67, 0.0122) (0.6900000000000001, 0.0002)
                (0.71, 0.0014) (0.73, 0.0001) (0.75, 0.0034) (0.77, 0.0) (0.79, 0.0001)
                (0.81, 0.0012) (0.8300000000000001, 0.0002) (0.85, 0.0) (0.87, 0.0) (0.89, 0.0002)
                (0.91, 0.0001) (0.93, 0.0) (0.95, 0.0) (0.97, 0.0) (0.99, 0.0021)
            };
    
            % \legend{min: 0.02, max: 1.00, median: 0.08, mean: 0.096}
            \end{axis}
        \end{tikzpicture}
        \caption{CMI Distribution}
        \label{fig:cmi}
    \end{subfigure} 
    
    \begin{subfigure}[b]{0.5\textwidth}
        % \centering
        \begin{tikzpicture}
            \begin{axis}[
                width=\textwidth, 
                height=3.8cm, 
                ylabel={Density (\%)}, 
                ylabel style={font=\small, yshift=-7pt}, 
                xmin=0, xmax=0.7, 
                ymin=0, ymax=10, 
                ytick={0, 5, 10},
                xtick pos=bottom,
                xticklabel style={font=\scriptsize, /pgf/number format/fixed}, 
                ytick pos=left,
                yticklabel style={font=\scriptsize, /pgf/number format/fixed}, 
                grid=major, 
                xmajorgrids=false,
                bar width=0.01,
                legend style={at={(0.98, 0.95)}, anchor=north east, font=\small}
            ]
            
            \addplot[ybar, fill=orange, opacity=0.7] coordinates {
                (0.01, 0.02871) (0.03, 0.264) (0.05, 1.6847) (0.07, 3.1214) (0.09, 4.1927)
                (0.11, 5.5454) (0.13, 4.0271) (0.15000000000000002, 4.2183) (0.16999999999999998, 3.4769) (0.19, 2.8312)
                (0.21000000000000002, 3.7192) (0.22999999999999998, 3.2499) (0.25, 2.1872) (0.27, 1.9621) (0.29000000000000004, 1.5036)
                (0.31, 1.7578) (0.33, 1.5351) (0.35, 0.4455) (0.37, 0.969) (0.39, 0.2021)
                (0.41000000000000003, 0.848) (0.43, 0.3555) (0.45, 0.4218) (0.47, 0.1643) (0.49, 0.0002)
                (0.51, 0.5699) (0.53, 0.0134) (0.55, 0.1571) (0.5700000000000001, 0.1255) (0.59, 0.0056)
                (0.61, 0.1139) (0.63, 0.0494) (0.65, 0.0013) (0.67, 0.1147) (0.6900000000000001, 0.0026)
                (0.71, 0.0292) (0.73, 0.0006) (0.75, 0.0386) (0.77, 0.0008) (0.79, 0.0004)
                (0.81, 0.0222) (0.8300000000000001, 0.0153) (0.85, 0.0064) (0.87, 0.0031) (0.89, 0.0)
                (0.91, 0.0) (0.93, 0.0) (0.95, 0.0) (0.97, 0.0209) (0.99, 0.0)
            };
    
            % \legend{Min: 0.00, Max: 1.00, Median: 0.14, Mean: 0.167}
            \end{axis}
        \end{tikzpicture}
        \caption{I-Index Distribution}
        \label{fig:i_index}
    \end{subfigure}
    
    \begin{subfigure}[b]{0.5\textwidth}
        % \centering
            \begin{tikzpicture}
                \begin{axis}[
                    width=\textwidth, 
                    height=3.8cm, 
                    ylabel={Density (\%)}, 
                    ylabel style={font=\small, yshift=-7pt}, 
                    xtick={1,2,...,26}, 
                    xticklabels={a, b, c, d, e, f, g, h, i, j, k, l, m, n, o, p, q, r, s, t, u, v, w, x, y, z},
                    x tick label style={rotate=0, font=\scriptsize, anchor=base, yshift=-7pt},
                    xtick pos=bottom,
                    grid=major, 
                    xmajorgrids=false,
                    legend style={at={(0.98, 0.95)}, column sep=5pt, font=\tiny, inner sep=1pt, legend cell align={left}, legend columns=1}, 
                    ymin=0, ymax=13,
                    ytick={0, 5, 10},
                    yticklabel style={font=\scriptsize, /pgf/number format/fixed}, 
                    xmin=0.25, xmax=26.75,
                    ybar,
                    % legend image code/.code={
                    %     \draw[fill=blue, opacity=0.5] (0,0) rectangle (0.05,0.05); 
                    %     \draw[fill=orange, opacity=0.5] (0,0.06) rectangle (0.05,0.11);
                    % },
                ]
                
                \addplot+[style={blue, opacity=0.5, bar width=0.25, xshift=0.12cm}] 
                    coordinates {(1, 11.309369) (2, 1.325023) (3, 2.275949) (4, 3.530119) (5, 12.367075)
                                (6, 0.930114) (7, 1.784955) (8, 1.813381) (9, 8.724409) (10, 1.337856)
                                (11, 2.638715) (12, 4.493302) (13, 2.96062) (14, 6.860761) (15, 7.059614)
                                (16, 2.482295) (17, 0.188604) (18, 5.593041) (19, 6.573522) (20, 6.754489)
                                (21, 3.702996) (22, 2.078873) (23, 0.512198) (24, 0.136589) (25, 0.998277)
                                (26, 1.567853)};
                
                \addplot+[style={orange, opacity=0.5, bar width=0.25, xshift=-0.12cm}] 
                    coordinates {(1, 11.349364) (2, 1.307629) (3, 2.296659) (4, 3.529734) (5, 12.41555)
                                (6, 0.930814) (7, 1.793685) (8, 1.811197) (9, 8.717389) (10, 1.26988)
                                (11, 2.61967) (12, 4.52354) (13, 2.954244) (14, 6.87295) (15, 7.060157)
                                (16, 2.464937) (17, 0.189312) (18, 5.585288) (19, 6.572185) (20, 6.761308)
                                (21, 3.699052) (22, 2.038446) (23, 0.514023) (24, 0.138004) (25, 1.012316)
                                (26, 1.572668)};
                
                \legend{CS-FLEURS, FLEURS-R}
                \end{axis}
            \end{tikzpicture}
        \caption{Romanized Character Diversity}
        \label{fig:phoneme_diversity}
    \end{subfigure}
    \caption{Data statistics of CS-FLEURS.}
    \label{fig:all_distributions}
% \vspace{-5pt}
\end{figure}
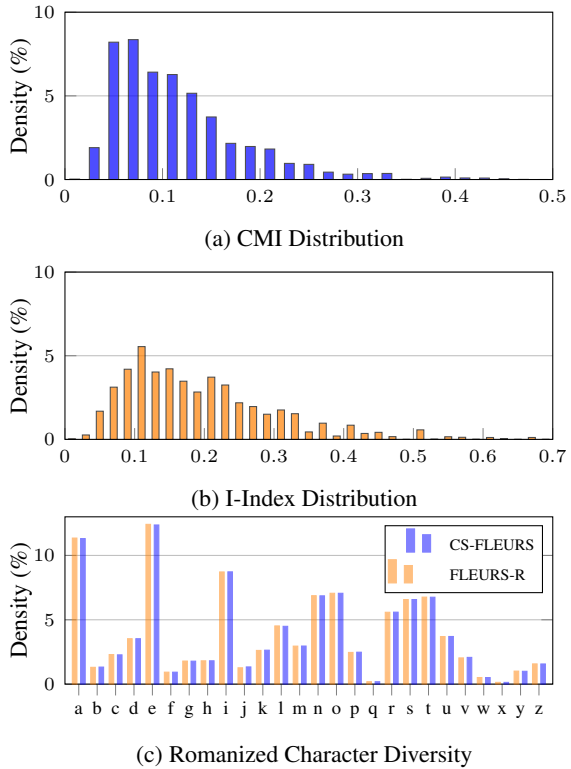

%% file: table/cs_fleurs_full.tex
\begin{table}[!t]
\centering
% \small
\resizebox{\columnwidth}{!}{%
\begin{tabular}{l|ccccc|c} \toprule
Pair     & it-es     & pt-es     & ro-es     & it-ro     & \multicolumn{1}{c}{it-pt} & de-es \\ \midrule
RER      &   20.5    &   22.8    &   23.0    &   23.1    &   \multicolumn{1}{c}{23.9}    &   24.8    \\
MOS      &    4.50   &   4.50    &    4.69   &   4.62    &    \multicolumn{1}{c}{4.89}   &    4.87   \\
SIS      &    4.42   &    4.85   &   4.71    &    4.65   &    \multicolumn{1}{c}{4.80}   &    4.94   \\ \midrule\midrule
\multicolumn{1}{c}{pl-es} & it-pl & cs-it & $\sim$ & bg-fi & \multicolumn{1}{l}{da-mt} & da-fi \\ \midrule
\multicolumn{1}{c}{24.9} &   25.1    &   25.2    &   $\sim$    &   38.1    &   \multicolumn{1}{c}{38.1}    &    38.3   \\
\multicolumn{1}{c}{4.28} &   4.75    &    4.32   &   $\sim$    &   4.84    &   \multicolumn{1}{c}{4.55}    &    4.21   \\
\multicolumn{1}{c}{4.29} &    4.76   &    5.00   &    $\sim$   &    4.86   &   \multicolumn{1}{c}{4.73}    &    4.53   \\ \midrule\midrule
\multicolumn{1}{c}{bg-el} & bg-en & hr-da & da-sl & da-el & bg-da & \textbf{Avg.} \\ \midrule
\multicolumn{1}{c}{38.7} &   38.7    &   38.9    &   39.3    &   40.2    &   41.9    &    \textbf{31.6}   \\
\multicolumn{1}{c}{4.36} &   4.05    &   4.71    &   4.59    &   4.00    &   4.60    & \textbf{4.44}      \\
\multicolumn{1}{c}{4.74} &   4.18    &   4.71    &   4.65    &   4.65    &   4.65    & \textbf{4.74}      \\ \bottomrule
\end{tabular}
}
\caption{Evaluation on the 9 best- and worst-performing language pairs of CS-FLEURS. \textit{Avg.} indicates the mean value across 253 pairs; full results are in Appendix~\ref{sec:lang_spec}.}
\label{tab_cs_fleurs_full}
\vspace{-5pt}
\end{table}

%% file: table/full_eval_merged.tex
\begin{table}[!t]
\small
\centering
\begin{subtable}[t]{0.48\textwidth}
\centering
\begin{tabular}{l|c|cc} \toprule
\multirow{2}{*}{Dataset} & Intelligibility & \multicolumn{2}{c}{Naturalness} \\
                         & RER $\downarrow$ & MOS $\uparrow$ & SIS $\uparrow$ \\ \midrule\midrule
SWC (de)                 & \textbf{25.8}    & \textbf{4.51}  & \textbf{4.90}  \\
MBC (es)                 & 56.9             & 4.20           & 4.40           \\ \midrule
CS-FLEURS (de)           & 30.1             & 4.36           & 4.83           \\ 
CS-FLEURS (es)           & 28.9             & 4.00           & 4.89           \\ \midrule
CS-FLEURS (avg)          & 31.6             & 4.44           & 4.74           \\ \bottomrule
\end{tabular}
\caption{In-domain (ID) language comparison.}
\label{tab_id}
\end{subtable}
\vspace{4pt}

\begin{subtable}[t]{0.48\textwidth}
\centering
\resizebox{\columnwidth}{!}{%
\begin{tabular}{l|c|cc} \toprule
\multirow{2}{*}{Dataset} & Intelligibility & \multicolumn{2}{c}{Naturalness} \\
                         & RER $\downarrow$ & MOS $\uparrow$ & SIS $\uparrow$ \\ \midrule\midrule
ASCEND (zh)              & -               & \textbf{4.85}  & 4.66           \\
MUCS2021 (hi)            & 48.3            & 4.50           & \textbf{4.87}  \\
MUCS2021 (bn)            & 57.6            & 3.87           & 4.28           \\ \midrule
CS-FLEURS-O (zh)         & 41.5            & 4.50           & 4.37           \\ 
CS-FLEURS-O (hi)         & 37.7            & 3.66           & 5.00           \\ 
CS-FLEURS-O (bn)         & 42.7            & 2.88           & 4.71           \\ \midrule
CS-FLEURS-O (avg)        & \textbf{40.8}    & 3.68           & 4.69           \\ \bottomrule
\end{tabular}
}
\caption{Out-of-domain (OOD) language comparison.}
\label{tab_ood}
\vspace{-5pt}
\end{subtable}
\caption{Quality comparison between CS-FLEURS and human-generated CS datasets.}
\label{tab:cs_fleurs_combined}
\vspace{-5pt}
\end{table}

%% file: table/impact_csasr.tex
\begin{table}[t!]
\centering
\small
\begin{tabular}{c|l|c|c} \toprule
Language Pair & Train Data & Eval Data & CER $\downarrow$ \\\midrule\midrule
\multirow{6}{*}{En-De}
  & SWC        & \multirow{3}{*}{SWC}        & 26.7 \\
  & CSF        &         & 45.8 \\
  & \textbf{SWC + CSF} &  & \textbf{23.0} \\\cmidrule{2-4}
  & SWC        &  \multirow{3}{*}{CSF}       & 48.3 \\
  & \textbf{CSF} &         & \textbf{23.7} \\
  & SWC + CSF  &         & 26.5 \\\midrule
\multirow{6}{*}{En-Es}
  & MBC        & \multirow{3}{*}{MBC}        & 100 \\
  & CSF        &         & 71.3 \\
  & \textbf{MBC + CSF} &  & \textbf{35.8} \\\cmidrule{2-4}
  & MBC        &  \multirow{3}{*}{CSF}       & 100 \\
  & \textbf{CSF} &         & \textbf{20.1} \\
  & MBC + CSF  &         & 22.9 \\\bottomrule
\end{tabular}
\caption{Impact of CS-FLEURS in CS-ASR training and evaluation for two in-domain language pairs.}
\label{tab_cs_impact}
\end{table}

%% file: table/appendix_prompt.tex
\begin{table*}[!t]
\centering
\small
\begin{tabular}{c|l|c} \toprule
 Type & \multicolumn{1}{c}{Detailed Prompt}               & Format                                                                         \\ \midrule\midrule
\begin{tabular}[c]{@{}c@{}}System\\(Role)\end{tabular} 
    & \begin{tabular}[c]{@{}l@{}}You are a language expert specializing in \textit{lang1} and \textit{lang2}.\end{tabular}     
    & \multirow{8}{*}{\begin{tabular}[c]{@{}l@{}}matches:\\ \qquad noun: \\ \qquad\qquad - [[n1, n2]]\\ \qquad verb: \\ \qquad\qquad - [[v1, v2]]\\ \qquad adverb: \\ \qquad\qquad - [[a1, a2]]\\ \qquad adjective: \\ \qquad\qquad - [[a'1, a'2]]\\ \qquad interjection: \\ \qquad\qquad - [[i1, i2]] 
\end{tabular}} 
    \\ \cmidrule(lr){1-2}
\begin{tabular}[c]{@{}c@{}}System\\(Formatting)\end{tabular} 
    & \renewcommand{\arraystretch}{1.5}\begin{tabular}[c]{@{}l@{}}The final outputs must be returned in YAML format,\\and each component in part of speech is a list of words with the same meaning.\\The YAML file structure must strictly adhere to the following format: \textit{format}\end{tabular} 
    &  
    \\ \cmidrule(lr){1-2}
\begin{tabular}[c]{@{}c@{}}User\\(Input)\end{tabular} 
    & \renewcommand{\arraystretch}{1.5}\begin{tabular}[c]{@{}l@{}}Find pairs of words with the same meaning and sort it with the part of speech\\information in the given two sentences from different languages.\\\textit{lang1} sentence: \textit{trans1}, \textit{lang2} sentence: \textit{trans2}\end{tabular} 
    &  
    \\ \bottomrule
\end{tabular}
\caption{Detailed prompting strategy of word-level mapping generation. Italicized text indicates the hyper-parameter of the pipeline.}
\label{prompts}
\end{table*}

%% file: figure/fig_wordmap.tex
\begin{figure*}[!t] 
    \centering
    \includegraphics[width=\textwidth]{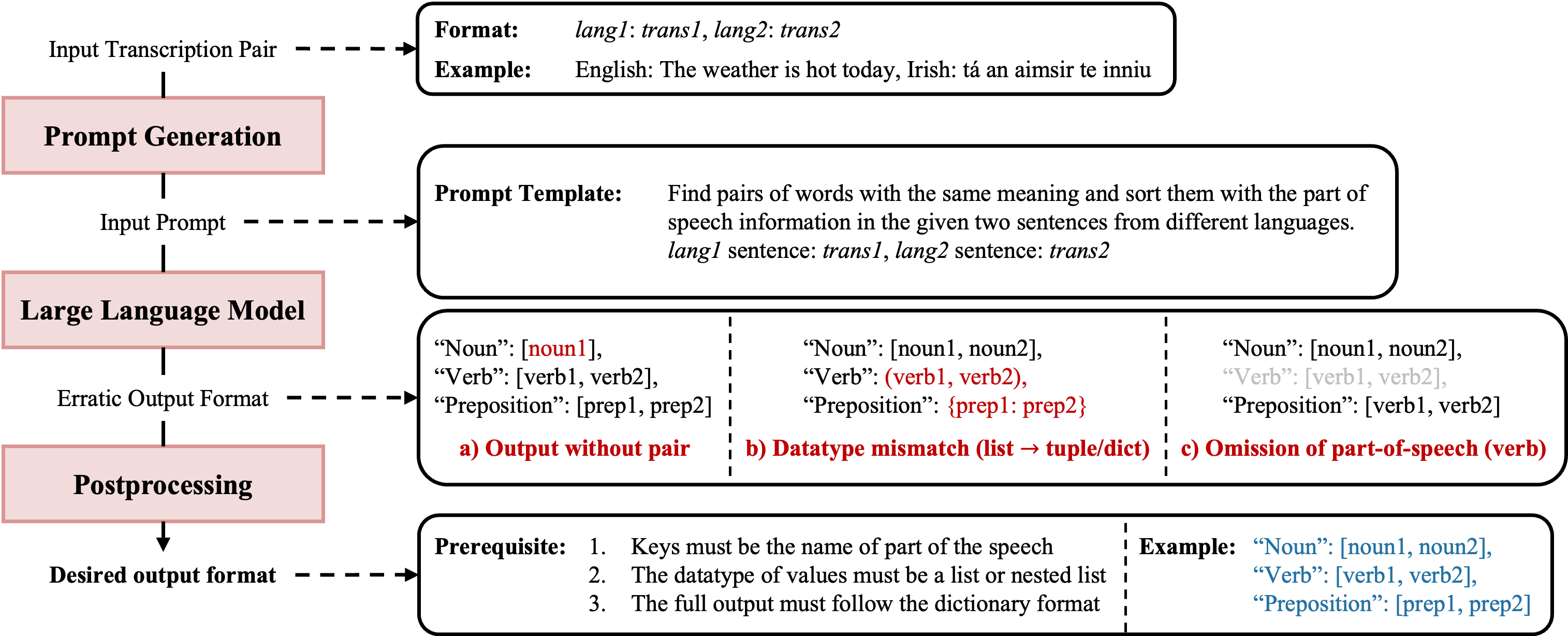}
    \caption{The overall pipeline of word-level mapping generation. Italicized text indicate hyper-parameter of the input prompt.}
    \label{fig:wordmap} 
\end{figure*}

%% file: table/appendix_full.tex
\begin{table*}[!ht]
\small
\centering
\begin{tabular}{l|lll||l|lll||l|lll} \toprule
Lang. pair & RER  & MOS  & SIS  & Lang. pair & RER  & MOS  & SIS  & Lang. pair & RER  & MOS  & SIS  \\ \midrule\midrule
bg-hr         & 36.1 & 3.94 & 4.44 & nl-fr         & 32.6 & 4.8  & 4.9  & de-ro         & 26.6 & 4.31 & 4.89 \\
bg-cs         & 32.9 & 4.53 & 4.61 & nl-de         & 27.8 & 4.53 & 4.95 & de-sk         & 29.9 & 4.41 & 4.94 \\
bg-da         & 41.9 & 4.6  & 4.65 & nl-el         & 36.0 & 4.63 & 4.93 & de-sl         & 31.2 & 4.6  & 4.89 \\
bg-nl         & 37.9 & 4.0  & 4.76 & nl-hu         & 32.7 & 4.8  & 4.85 & de-es         & 24.8 & 4.87 & 4.94 \\
bg-en         & 38.7 & 4.05 & 4.18 & nl-it         & 29.1 & 4.67 & 4.62 & de-sv         & 28.4 & 4.54 & 4.6  \\
bg-et         & 37.5 & 4.0  & 4.56 & nl-lv         & 32.5 & 4.75 & 4.62 & el-hu         & 34.0 & 4.38 & 4.58 \\
bg-fi         & 38.1 & 4.84 & 4.86 & nl-lt         & 33.4 & 4.44 & 4.84 & el-it         & 29.9 & 4.6  & 4.94 \\
bg-fr         & 37.0 & 3.5  & 4.58 & nl-mt         & 34.3 & 4.75 & 4.83 & el-lv         & 33.4 & 4.44 & 4.76 \\
bg-de         & 34.6 & 4.0  & 4.69 & nl-pl         & 31.4 & 4.44 & 4.8  & el-lt         & 33.5 & 4.62 & 4.94 \\
bg-el         & 38.7 & 4.36 & 4.74 & nl-pt         & 31.7 & 4.44 & 4.64 & el-mt         & 35.5 & 4.5  & 4.78 \\
bg-hu         & 35.9 & 3.67 & 4.41 & nl-ro         & 30.7 & 4.79 & 4.93 & el-pl         & 32.3 & 4.81 & 4.88 \\
bg-it         & 31.4 & 4.06 & 4.78 & nl-sk         & 33.3 & 4.44 & 4.63 & el-pt         & 32.3 & 4.0  & 4.75 \\
bg-lv         & 34.3 & 3.94 & 4.75 & nl-sl         & 35.0 & 4.55 & 4.61 & el-ro         & 31.1 & 4.43 & 4.89 \\
bg-lt         & 34.9 & 4.28 & 4.94 & nl-es         & 28.7 & 4.43 & 4.71 & el-sk         & 34.0 & 4.69 & 5.0  \\
bg-mt         & 37.8 & 3.44 & 4.72 & nl-sv         & 32.6 & 4.53 & 4.38 & el-sl         & 35.7 & 4.81 & 4.88 \\
bg-pl         & 33.3 & 4.88 & 4.89 & en-et         & 35.6 & 4.78 & 4.73 & el-es         & 29.5 & 4.44 & 4.81 \\
bg-pt         & 33.8 & 4.0  & 4.73 & en-fi         & 35.9 & 4.57 & 4.88 & el-sv         & 35.2 & 4.47 & 4.5  \\
bg-ro         & 31.9 & 3.95 & 4.84 & en-fr         & 32.9 & 3.94 & 4.47 & hu-it         & 27.7 & 4.35 & 4.79 \\
bg-sk         & 34.1 & 4.53 & 4.88 & en-de         & 30.1 & 4.36 & 4.83 & hu-lv         & 30.1 & 4.3  & 4.8  \\ 
bg-sl         & 35.9 & 4.28 & 4.75 & en-el         & 36.6 & 4.5  & 4.88 & hu-lt         & 31.6 & 4.41 & 4.95 \\ 
bg-es         & 31.8 & 4.23 & 4.47 & en-hu         & 33.5 & 4.67 & 4.73 & hu-mt         & 32.2 & 4.88 & 5.0  \\
bg-sv         & 37.3 & 4.07 & 4.65 & en-it         & 29.3 & 4.16 & 4.53 & hu-pl         & 29.7 & 4.61 & 4.83 \\ \bottomrule
\end{tabular}
\caption{Full experimental results for 253 language pairs. All the languages are denoted with ISO-639-1 code.}
\label{appendix_full_pt1}
\end{table*}

%% file: table/appendix_full2.tex
\begin{table*}[!ht]
\small
\centering
\begin{tabular}{l|lll||l|lll||l|lll} \toprule
Lang. pair & RER  & MOS  & SIS  & Lang. pair & RER  & MOS  & SIS  & Lang. pair & RER  & MOS  & SIS  \\ \midrule\midrule
hr-cs         & 29.2 & 4.29 & 4.75 & en-lv         & 32.5 & 4.0  & 4.81 & hu-pt         & 30.2 & 4.4  & 4.75 \\
hr-da         & 38.9 & 4.71 & 4.71 & en-lt         & 33.5 & 4.0  & 4.82 & hu-ro         & 29.2 & 4.88 & 4.8  \\
hr-nl         & 34.7 & 4.12 & 4.79 & en-mt         & 34.5 & 4.35 & 4.72 & hu-sk         & 30.9 & 4.74 & 4.94 \\
hr-en         & 35.1 & 4.75 & 4.7  & en-pl         & 32.1 & 4.5  & 4.63 & hu-sl         & 32.5 & 4.5  & 4.94 \\
hr-et         & 34.2 & 4.44 & 4.85 & en-pt         & 30.8 & 3.93 & 4.44 & hu-es         & 27.2 & 4.59 & 4.94 \\
hr-fi         & 34.7 & 4.36 & 4.78 & en-ro         & 30.9 & 4.47 & 4.73 & hu-sv         & 32.1 & 4.57 & 4.84 \\
hr-fr         & 34.4 & 4.19 & 4.13 & en-sk         & 33.8 & 4.5  & 4.5  & it-lv         & 26.0 & 4.67 & 4.67 \\
hr-de         & 31.1 & 4.15 & 4.76 & en-sl         & 35.7 & 4.65 & 4.87 & it-lt         & 26.6 & 4.89 & 4.61 \\
hr-el         & 35.6 & 4.57 & 4.89 & en-es         & 28.9 & 4.0  & 4.89 & it-mt         & 27.4 & 4.45 & 4.58 \\
hr-hu         & 32.2 & 4.56 & 4.44 & en-sv         & 33.7 & 4.24 & 4.88 & it-pl         & 25.1 & 4.75 & 4.76 \\
hr-it         & 28.3 & 4.18 & 4.9  & et-fi         & 31.6 & 4.61 & 4.94 & it-pt         & 23.9 & 4.89 & 4.8  \\
hr-lv         & 30.6 & 4.35 & 4.83 & et-fr         & 34.3 & 4.59 & 4.86 & it-ro         & 23.1 & 4.62 & 4.65 \\
hr-lt         & 31.5 & 4.65 & 4.8  & et-de         & 30.7 & 4.44 & 4.88 & it-sk         & 26.4 & 4.65 & 4.95 \\
hr-mt         & 33.7 & 4.37 & 4.59 & et-el         & 35.9 & 4.61 & 4.79 & it-sl         & 28.4 & 4.78 & 4.95 \\
hr-pl         & 29.3 & 4.36 & 4.69 & et-hu         & 32.0 & 4.29 & 4.93 & it-es         & 20.5 & 4.5  & 4.42 \\
hr-pt         & 30.7 & 4.53 & 4.5  & et-it         & 28.8 & 4.84 & 4.94 & it-sv         & 27.9 & 3.78 & 4.68 \\
hr-ro         & 28.9 & 4.33 & 4.88 & et-lv         & 30.7 & 4.53 & 4.88 & lv-lt         & 28.4 & 4.4  & 4.62 \\
hr-sk         & 30.3 & 4.69 & 4.69 & et-lt         & 32.4 & 4.56 & 4.86 & lv-mt         & 31.2 & 4.17 & 4.69 \\
hr-sl         & 31.7 & 4.62 & 4.4  & et-mt         & 33.5 & 4.39 & 4.65 & lv-pl         & 28.8 & 4.33 & 4.56 \\
hr-es         & 27.7 & 4.62 & 4.61 & et-pl         & 31.3 & 4.54 & 4.78 & lv-pt         & 28.3 & 4.53 & 4.44 \\
hr-sv         & 33.4 & 4.37 & 4.65 & et-pt         & 31.6 & 4.31 & 4.72 & lv-ro         & 28.1 & 4.11 & 4.76 \\
cs-da         & 35.8 & 4.26 & 4.36 & et-ro         & 30.4 & 4.73 & 4.93 & lv-sk         & 28.8 & 4.72 & 4.81 \\
cs-nl         & 31.3 & 4.65 & 4.47 & et-sk         & 32.3 & 4.11 & 4.75 & lv-sl         & 31.5 & 4.5  & 4.84 \\
cs-en         & 31.4 & 4.37 & 4.7  & et-sl         & 34.5 & 4.63 & 4.78 & lv-es         & 26.3 & 4.06 & 4.82 \\
cs-et         & 31.2 & 3.65 & 4.5  & et-es         & 28.2 & 4.65 & 4.78 & lv-sv         & 30.9 & 4.25 & 4.57 \\
cs-fi         & 31.7 & 4.77 & 4.88 & et-sv         & 33.1 & 3.95 & 4.8  & lt-mt         & 32.1 & 4.81 & 4.67 \\
cs-fr         & 30.6 & 4.0  & 4.93 & fi-fr         & 35.4 & 4.18 & 4.88 & lt-pl         & 29.3 & 4.75 & 4.63 \\
cs-de         & 28.1 & 4.5  & 4.75 & fi-de         & 31.0 & 3.94 & 4.47 & lt-pt         & 29.1 & 4.56 & 4.81 \\
cs-el         & 33.2 & 4.29 & 4.63 & fi-el         & 36.2 & 4.27 & 4.71 & lt-ro         & 28.3 & 4.72 & 4.76 \\
cs-hu         & 29.4 & 4.92 & 4.75 & fi-hu         & 32.5 & 4.32 & 4.9  & lt-sk         & 30.4 & 4.68 & 4.9  \\
cs-it         & 25.2 & 4.32 & 5.0  & fi-it         & 29.4 & 4.67 & 4.94 & lt-sl         & 32.2 & 4.59 & 4.86 \\
cs-lv         & 28.1 & 4.65 & 5.0  & fi-lv         & 31.2 & 4.17 & 4.72 & lt-es         & 26.3 & 3.9  & 4.78 \\
cs-lt         & 29.9 & 4.44 & 5.0  & fi-lt         & 32.5 & 4.38 & 4.47 & lt-sv         & 32.3 & 4.69 & 4.62 \\
cs-mt         & 30.6 & 4.41 & 4.67 & fi-mt         & 33.9 & 4.28 & 4.82 & mt-pl         & 30.5 & 4.47 & 4.56 \\
cs-pl         & 25.7 & 4.53 & 4.94 & fi-pl         & 32.1 & 4.47 & 4.73 & mt-pt         & 30.6 & 4.35 & 4.77 \\
cs-pt         & 27.8 & 3.94 & 4.61 & fi-pt         & 32.6 & 4.06 & 4.8  & mt-ro         & 28.9 & 4.53 & 4.71 \\
cs-ro         & 26.5 & 4.41 & 4.77 & fi-ro         & 31.5 & 4.9  & 4.87 & mt-sk         & 31.7 & 4.81 & 4.86 \\
cs-sk         & 26.1 & 4.63 & 4.76 & fi-sk         & 33.0 & 4.44 & 4.83 & mt-sl         & 33.9 & 4.8  & 4.76 \\
cs-sl         & 29.4 & 4.74 & 4.71 & fi-sl         & 35.3 & 4.44 & 5.0  & mt-es         & 27.8 & 4.5  & 4.65 \\
cs-es         & 25.3 & 4.63 & 4.82 & fi-es         & 28.9 & 4.41 & 4.79 & mt-sv         & 32.9 & 4.59 & 4.74 \\
cs-sv         & 30.8 & 3.9  & 4.81 & fi-sv         & 33.7 & 4.13 & 4.56 & pl-pt         & 28.2 & 4.35 & 4.58 \\
da-nl         & 36.2 & 4.58 & 4.59 & fr-de         & 29.1 & 4.72 & 4.45 & pl-ro         & 26.6 & 4.62 & 4.94 \\
da-en         & 37.3 & 4.12 & 4.89 & fr-el         & 35.2 & 4.33 & 4.75 & pl-sk         & 27.3 & 4.63 & 4.62 \\
da-et         & 38.0 & 4.88 & 4.45 & fr-hu         & 33.0 & 4.33 & 4.61 & pl-sl         & 29.5 & 4.62 & 4.79 \\
da-fi         & 38.3 & 4.21 & 4.53 & fr-it         & 27.2 & 4.65 & 4.63 & pl-es         & 24.9 & 4.28 & 4.29 \\
da-fr         & 36.8 & 4.53 & 4.88 & fr-lv         & 32.1 & 3.88 & 4.75 & pl-sv         & 31.1 & 4.56 & 4.71 \\
da-de         & 31.9 & 4.42 & 4.88 & fr-lt         & 33.7 & 4.18 & 4.59 & pt-ro         & 25.4 & 4.13 & 4.85 \\
da-el         & 40.2 & 4.0  & 4.65 & fr-mt         & 34.1 & 4.2  & 4.69 & pt-sk         & 29.4 & 4.4  & 5.0  \\
da-hu         & 37.3 & 4.47 & 4.89 & fr-pl         & 30.5 & 4.72 & 4.61 & pt-sl         & 31.4 & 4.19 & 4.38 \\
da-it         & 33.0 & 4.65 & 4.24 & fr-pt         & 28.8 & 4.2  & 4.57 & pt-es         & 22.8 & 4.5  & 4.85 \\
da-lv         & 36.7 & 4.37 & 5.0  & fr-ro         & 29.0 & 4.25 & 4.8  & pt-sv         & 30.5 & 4.25 & 4.81 \\
da-lt         & 37.8 & 4.39 & 4.56 & fr-sk         & 33.2 & 4.74 & 4.41 & ro-sk         & 29.0 & 4.75 & 4.83 \\
da-mt         & 38.1 & 4.55 & 4.73 & fr-sl         & 34.4 & 4.22 & 4.89 & ro-sl         & 29.7 & 4.74 & 4.89 \\
da-pl         & 36.0 & 4.24 & 5.0  & fr-es         & 26.8 & 4.5  & 4.68 & ro-es         & 23.0 & 4.69 & 4.71 \\
da-pt         & 35.2 & 4.28 & 4.87 & fr-sv         & 33.1 & 3.87 & 4.74 & ro-sv         & 30.2 & 4.53 & 4.68 \\
da-ro         & 34.4 & 4.89 & 4.83 & de-el         & 32.9 & 4.4  & 4.71 & sk-sl         & 30.4 & 4.61 & 4.65 \\
da-sk         & 37.3 & 4.56 & 4.87 & de-hu         & 29.4 & 4.5  & 5.0  & sk-es         & 26.5 & 4.5  & 4.71 \\
da-sl         & 39.3 & 4.59 & 4.65 & de-it         & 25.5 & 4.53 & 4.86 & sk-sv         & 32.1 & 4.78 & 4.74 \\
da-es         & 33.0 & 4.83 & 4.11 & de-lv         & 28.7 & 4.35 & 4.63 & sl-es         & 28.3 & 4.89 & 4.94 \\
da-sv         & 35.7 & 4.35 & 4.71 & de-lt         & 30.0 & 4.5  & 5.0  & sl-sv         & 34.2 & 4.53 & 4.72 \\
nl-en         & 33.6 & 4.71 & 4.7  & de-mt         & 30.8 & 4.39 & 4.89 & es-sv         & 28.1 & 4.0  & 4.83 \\
nl-et         & 34.7 & 4.62 & 4.74 & de-pl         & 28.1 & 4.44 & 4.8  & \textbf{Avg.}       & \textbf{31.6} & \textbf{4.44} & \textbf{4.74} \\
nl-fi         & 35.1 & 4.32 & 4.71 & de-pt         & 27.7 & 4.32 & 4.75 &               &      &      &      \\ \bottomrule
\end{tabular}
\caption{Full experimental results for 253 language pairs. All the languages are denoted with ISO-639-1 code.}
\label{appendix_full_pt2}
\end{table*}

%% file: figure/appendix_distribution.tex
\begin{figure*}[!t]
    \centering
    \begin{subfigure}[b]{0.9\textwidth}
        % \centering
        \begin{tikzpicture}
            \begin{axis}[
                width=\textwidth, 
                height=5.5cm, 
                ylabel={RER $\downarrow$}, 
                ylabel style={font=\small, yshift=-7pt}, 
                xlabel={Language pair index}, 
                xlabel style={font=\small, yshift=3pt}, 
                xmin=0, xmax=253, 
                ymin=20, ymax=60, 
                ytick={20, 30, 40, 50, 60},
                xtick pos=bottom,
                xticklabel style={font=\small, /pgf/number format/fixed}, 
                ytick pos=left,
                yticklabel style={font=\small, /pgf/number format/fixed}, 
                grid=major, 
                xmajorgrids=false,
                bar width=0.1,
                legend style={at={(0.98, 0.95)}, anchor=north east, font=\small}
            ]
            
            \addplot[ybar, fill=blue, opacity=1, draw=none] coordinates {
                (1,20.5) (2,22.8) (3,23.0) (4,23.1) (5,23.9) (6,24.8) (7,24.9) (8,25.1) (9,25.2) (10,25.3)
                (11,25.4) (12,25.5) (13,25.7) (14,26.0) (15,26.1) (16,26.3) (17,26.3) (18,26.4) (19,26.5) (20,26.5)
                (21,26.6) (22,26.6) (23,26.6) (24,26.8) (25,27.2) (26,27.2) (27,27.3) (28,27.4) (29,27.7) (30,27.7)
                (31,27.7) (32,27.8) (33,27.8) (34,27.8) (35,27.9) (36,28.1) (37,28.1) (38,28.1) (39,28.1) (40,28.1)
                (41,28.2) (42,28.2) (43,28.3) (44,28.3) (45,28.3) (46,28.3) (47,28.4) (48,28.4) (49,28.4) (50,28.7)
                (51,28.7) (52,28.8) (53,28.8) (54,28.8) (55,28.8) (56,28.9) (57,28.9) (58,28.9) (59,28.9) (60,29.0)
                (61,29.0) (62,29.1) (63,29.1) (64,29.1) (65,29.2) (66,29.2) (67,29.3) (68,29.3) (69,29.3) (70,29.4)
                (71,29.4) (72,29.4) (73,29.4) (74,29.4) (75,29.5) (76,29.5) (77,29.7) (78,29.7) (79,29.9) (80,29.9)
                (81,29.9) (82,30.0) (83,30.1) (84,30.1) (85,30.2) (86,30.2) (87,30.3) (88,30.4) (89,30.4) (90,30.4)
                (91,30.5) (92,30.5) (93,30.5) (94,30.6) (95,30.6) (96,30.6) (97,30.6) (98,30.7) (99,30.7) (100,30.7)
                (101,30.7) (102,30.8) (103,30.8) (104,30.8) (105,30.9) (106,30.9) (107,30.9) (108,31.0) (109,31.1)
                (110,31.1) (111,31.1) (112,31.2) (113,31.2) (114,31.2) (115,31.2) (116,31.3) (117,31.3) (118,31.4)
                (119,31.4) (120,31.4) (121,31.4) (122,31.5) (123,31.5) (124,31.5) (125,31.6) (126,31.6) (127,31.6)
                (128,31.6) (129,31.7) (130,31.7) (131,31.7) (132,31.7) (133,31.8) (134,31.9) (135,31.9) (136,32.0)
                (137,32.1) (138,32.1) (139,32.1) (140,32.1) (141,32.1) (142,32.1) (143,32.2) (144,32.2) (145,32.2)
                (146,32.3) (147,32.3) (148,32.3) (149,32.3) (150,32.4) (151,32.5) (152,32.5) (153,32.5) (154,32.5)
                (155,32.5) (156,32.6) (157,32.6) (158,32.6) (159,32.7) (160,32.9) (161,32.9) (162,32.9) (163,32.9)
                (164,33.0) (165,33.0) (166,33.0) (167,33.0) (168,33.1) (169,33.1) (170,33.2) (171,33.2) (172,33.3)
                (173,33.3) (174,33.4) (175,33.4) (176,33.4) (177,33.5) (178,33.5) (179,33.5) (180,33.5) (181,33.6)
                (182,33.7) (183,33.7) (184,33.7) (185,33.7) (186,33.8) (187,33.8) (188,33.9) (189,33.9) (190,34.0)
                (191,34.0) (192,34.1) (193,34.1) (194,34.2) (195,34.2) (196,34.3) (197,34.3) (198,34.3) (199,34.4)
                (200,34.4) (201,34.4) (202,34.5) (203,34.5) (204,34.6) (205,34.7) (206,34.7) (207,34.7) (208,34.9)
                (209,35.0) (210,35.1) (211,35.1) (212,35.2) (213,35.2) (214,35.2) (215,35.3) (216,35.4) (217,35.5)
                (218,35.6) (219,35.6) (220,35.7) (221,35.7) (222,35.7) (223,35.8) (224,35.9) (225,35.9) (226,35.9)
                (227,35.9) (228,36.0) (229,36.0) (230,36.1) (231,36.2) (232,36.2) (233,36.6) (234,36.7) (235,36.8)
                (236,37.0) (237,37.3) (238,37.3) (239,37.3) (240,37.3) (241,37.5) (242,37.8) (243,37.8) (244,37.9)
                (245,38.0) (246,38.1) (247,38.1) (248,38.3) (249,38.7) (250,38.7) (251,38.9) (252,39.3) (253,40.2)
            };
            
            \draw[red, thick] (axis cs:0,41.4) -- (axis cs:253,41.4);
            \draw[red, thick, dashed] (axis cs:0,53.0) -- (axis cs:253,53.0);
            \draw[blue, thick] (axis cs:0,31.6) -- (axis cs:253,31.6);
            % \legend{Min: 0.00, Max: 1.00, Median: 0.14, Mean: 0.167}
            \end{axis}
        \end{tikzpicture}
        \caption{RER Distribution}
        \label{fig:rcer}
    \end{subfigure} 
    
    \begin{subfigure}[b]{0.9\textwidth}
        % \centering
        \begin{tikzpicture}
            \begin{axis}[
                width=\textwidth, 
                height=5.5cm, 
                ylabel={MOS $\uparrow$}, 
                ylabel style={font=\small, yshift=-7pt}, 
                xlabel={Language pair index}, 
                xlabel style={font=\small, yshift=3pt}, 
                xmin=0, xmax=253, 
                ymin=3, ymax=5, 
                ytick={3, 3.5, 4, 4.5, 5},
                xtick pos=bottom,
                xticklabel style={font=\small, /pgf/number format/fixed}, 
                ytick pos=left,
                yticklabel style={font=\small, /pgf/number format/fixed}, 
                grid=major, 
                xmajorgrids=false,
                bar width=0.1,
                legend style={at={(0.98, 0.95)}, anchor=north east, font=\small}
            ]
            
            \addplot[ybar, fill=orange, opacity=1, draw=none] coordinates {
                (1,3.44) (2,3.5) (3,3.65) (4,3.67) (5,3.78) (6,3.87) (7,3.88) (8,3.9) (9,3.9) (10,3.93)
                (11,3.94) (12,3.94) (13,3.94) (14,3.94) (15,3.94) (16,3.95) (17,3.95) (18,4.0) (19,4.0) (20,4.0)
                (21,4.0) (22,4.0) (23,4.0) (24,4.0) (25,4.0) (26,4.0) (27,4.0) (28,4.0) (29,4.0) (30,4.0)
                (31,4.0) (32,4.05) (33,4.06) (34,4.06) (35,4.06) (36,4.07) (37,4.11) (38,4.11) (39,4.12) (40,4.12)
                (41,4.13) (42,4.13) (43,4.15) (44,4.16) (45,4.17) (46,4.17) (47,4.18) (48,4.18) (49,4.18) (50,4.19)
                (51,4.19) (52,4.2) (53,4.2) (54,4.21) (55,4.22) (56,4.23) (57,4.24) (58,4.24) (59,4.25) (60,4.25)
                (61,4.25) (62,4.26) (63,4.27) (64,4.28) (65,4.28) (66,4.28) (67,4.28) (68,4.28) (69,4.29) (70,4.29)
                (71,4.29) (72,4.3) (73,4.31) (74,4.31) (75,4.32) (76,4.32) (77,4.32) (78,4.32) (79,4.33) (80,4.33)
                (81,4.33) (82,4.33) (83,4.35) (84,4.35) (85,4.35) (86,4.35) (87,4.35) (88,4.35) (89,4.35) (90,4.36)
                (91,4.36) (92,4.36) (93,4.36) (94,4.37) (95,4.37) (96,4.37) (97,4.37) (98,4.38) (99,4.38) (100,4.39)
                (101,4.39) (102,4.39) (103,4.4) (104,4.4) (105,4.4) (106,4.4) (107,4.41) (108,4.41) (109,4.41) (110,4.41)
                (111,4.41) (112,4.42) (113,4.43) (114,4.43) (115,4.44) (116,4.44) (117,4.44) (118,4.44) (119,4.44) (120,4.44)
                (121,4.44) (122,4.44) (123,4.44) (124,4.44) (125,4.44) (126,4.44) (127,4.45) (128,4.47) (129,4.47) (130,4.47)
                (131,4.47) (132,4.47) (133,4.5) (134,4.5) (135,4.5) (136,4.5) (137,4.5) (138,4.5) (139,4.5) (140,4.5)
                (141,4.5) (142,4.5) (143,4.5) (144,4.5) (145,4.5) (146,4.5) (147,4.53) (148,4.53) (149,4.53) (150,4.53)
                (151,4.53) (152,4.53) (153,4.53) (154,4.53) (155,4.53) (156,4.53) (157,4.53) (158,4.53) (159,4.54) (160,4.54)
                (161,4.55) (162,4.55) (163,4.56) (164,4.56) (165,4.56) (166,4.56) (167,4.56) (168,4.57) (169,4.57) (170,4.57)
                (171,4.58) (172,4.59) (173,4.59) (174,4.59) (175,4.59) (176,4.59) (177,4.6) (178,4.6) (179,4.6) (180,4.61)
                (181,4.61) (182,4.61) (183,4.61) (184,4.62) (185,4.62) (186,4.62) (187,4.62) (188,4.62) (189,4.62) (190,4.62)
                (191,4.63) (192,4.63) (193,4.63) (194,4.63) (195,4.63) (196,4.65) (197,4.65) (198,4.65) (199,4.65) (200,4.65)
                (201,4.65) (202,4.65) (203,4.67) (204,4.67) (205,4.67) (206,4.67) (207,4.68) (208,4.69) (209,4.69) (210,4.69)
                (211,4.69) (212,4.71) (213,4.71) (214,4.72) (215,4.72) (216,4.72) (217,4.72) (218,4.73) (219,4.74) (220,4.74)
                (221,4.74) (222,4.74) (223,4.75) (224,4.75) (225,4.75) (226,4.75) (227,4.75) (228,4.75) (229,4.77) (230,4.78)
                (231,4.78) (232,4.78) (233,4.79) (234,4.8) (235,4.8) (236,4.8) (237,4.81) (238,4.81) (239,4.81) (240,4.81)
                (241,4.83) (242,4.84) (243,4.84) (244,4.87) (245,4.88) (246,4.88) (247,4.88) (248,4.88) (249,4.89) (250,4.89)
                (251,4.89) (252,4.9) (253,4.92)
            };
            \draw[red, thick] (axis cs:0,4.35) -- (axis cs:253,4.35);
            \draw[red, thick, dashed] (axis cs:0,4.41) -- (axis cs:253,4.41);
            \draw[blue, thick] (axis cs:0,4.44) -- (axis cs:253,4.44);
            % \legend{Min: 0.00, Max: 1.00, Median: 0.14, Mean: 0.167}
            \end{axis}
        \end{tikzpicture}
        \caption{MOS Distribution}
        \label{fig:mos}
    \end{subfigure} 
    
    \begin{subfigure}[b]{0.9\textwidth}
        % \centering
        \begin{tikzpicture}
            \begin{axis}[
                width=\textwidth, 
                height=5.5cm, 
                ylabel={SIS $\uparrow$}, 
                ylabel style={font=\small, yshift=-7pt}, 
                xlabel={Language pair index}, 
                xlabel style={font=\small, yshift=3pt}, 
                xmin=0, xmax=253, 
                ymin=3, ymax=5, 
                ytick={3, 3.5, 4, 4.5, 5},
                xtick pos=bottom,
                xticklabel style={font=\small, /pgf/number format/fixed}, 
                ytick pos=left,
                yticklabel style={font=\small, /pgf/number format/fixed}, 
                grid=major, 
                xmajorgrids=false,
                bar width=0.1,
                legend style={at={(0.98, 0.95)}, anchor=north east, font=\small}
            ]
            
            \addplot[ybar, fill=orange, opacity=1, draw=none] coordinates {
                (1,4.11) (2,4.13) (3,4.18) (4,4.24) (5,4.29) (6,4.36) (7,4.38) (8,4.38) (9,4.4) (10,4.41)
                (11,4.41) (12,4.42) (13,4.44) (14,4.44) (15,4.44) (16,4.44) (17,4.45) (18,4.45) (19,4.47) (20,4.47)
                (21,4.47) (22,4.47) (23,4.47) (24,4.5) (25,4.5) (26,4.5) (27,4.5) (28,4.53) (29,4.53) (30,4.56) (31,4.56)
                (32,4.56) (33,4.56) (34,4.56) (35,4.57) (36,4.57) (37,4.58) (38,4.58) (39,4.58) (40,4.58) (41,4.59) (42,4.59)
                (43,4.59) (44,4.6) (45,4.61) (46,4.61) (47,4.61) (48,4.61) (49,4.61) (50,4.61) (51,4.61) (52,4.62) (53,4.62)
                (54,4.62) (55,4.62) (56,4.62) (57,4.63) (58,4.63) (59,4.63) (60,4.63) (61,4.63) (62,4.63) (63,4.64) (64,4.65)
                (65,4.65) (66,4.65) (67,4.65) (68,4.65) (69,4.65) (70,4.65) (71,4.65) (72,4.65) (73,4.67) (74,4.67) (75,4.67)
                (76,4.68) (77,4.68) (78,4.68) (79,4.69) (80,4.69) (81,4.69) (82,4.69) (83,4.69) (84,4.7) (85,4.7) (86,4.7)
                (87,4.71) (88,4.71) (89,4.71) (90,4.71) (91,4.71) (92,4.71) (93,4.71) (94,4.71) (95,4.71) (96,4.71) (97,4.71)
                (98,4.72) (99,4.72) (100,4.72) (101,4.72) (102,4.72) (103,4.73) (104,4.73) (105,4.73) (106,4.73) (107,4.73) (108,4.73)
                (109,4.73) (110,4.74) (111,4.74) (112,4.74) (113,4.74) (114,4.74) (115,4.74) (116,4.75) (117,4.75) (118,4.75) (119,4.75)
                (120,4.75) (121,4.75) (122,4.75) (123,4.75) (124,4.75) (125,4.75) (126,4.75) (127,4.76) (128,4.76) (129,4.76) (130,4.76)
                (131,4.76) (132,4.76) (133,4.76) (134,4.76) (135,4.77) (136,4.77) (137,4.78) (138,4.78) (139,4.78) (140,4.78) (141,4.78)
                (142,4.78) (143,4.78) (144,4.78) (145,4.79) (146,4.79) (147,4.79) (148,4.79) (149,4.79) (150,4.8) (151,4.8) (152,4.8)
                (153,4.8) (154,4.8) (155,4.8) (156,4.8) (157,4.8) (158,4.8) (159,4.81) (160,4.81) (161,4.81) (162,4.81) (163,4.81)
                (164,4.81) (165,4.82) (166,4.82) (167,4.82) (168,4.82) (169,4.83) (170,4.83) (171,4.83) (172,4.83) (173,4.83) (174,4.83)
                (175,4.83) (176,4.83) (177,4.84) (178,4.84) (179,4.84) (180,4.84) (181,4.85) (182,4.85) (183,4.85) (184,4.85) (185,4.86)
                (186,4.86) (187,4.86) (188,4.86) (189,4.86) (190,4.86) (191,4.87) (192,4.87) (193,4.87) (194,4.87) (195,4.88) (196,4.88)
                (197,4.88) (198,4.88) (199,4.88) (200,4.88) (201,4.88) (202,4.88) (203,4.88) (204,4.88) (205,4.88) (206,4.88) (207,4.88)
                (208,4.88) (209,4.89) (210,4.89) (211,4.89) (212,4.89) (213,4.89) (214,4.89) (215,4.89) (216,4.89) (217,4.89) (218,4.89)
                (219,4.89) (220,4.9) (221,4.9) (222,4.9) (223,4.9) (224,4.93) (225,4.93) (226,4.93) (227,4.93) (228,4.93) (229,4.94)
                (230,4.94) (231,4.94) (232,4.94) (233,4.94) (234,4.94) (235,4.94) (236,4.94) (237,4.94) (238,4.94) (239,4.94) (240,4.94)
                (241,4.94) (242,4.95) (243,4.95) (244,4.95) (245,4.95) (246,5.0) (247,5.0) (248,5.0) (249,5.0) (250,5.0) (251,5.0)
                (252,5.0)
            };
            
            \draw[red, thick] (axis cs:0,4.65) -- (axis cs:253,4.65);
            \draw[red, thick, dashed] (axis cs:0,4.60) -- (axis cs:253,4.60);
            \draw[blue, thick] (axis cs:0,4.74) -- (axis cs:253,4.74);
            % \legend{Min: 0.00, Max: 1.00, Median: 0.14, Mean: 0.167}
            \end{axis}
        \end{tikzpicture}
        \caption{SIS Distribution}
        \label{fig:rcer}
    \end{subfigure} 
    \caption{Performance distribution of CS-FLEURS. The blue line indicates the average for CS-FLEURS. Solid and dashed red lines represent the average for human-generated in-domain and out-of-domain datasets, respectively.}
    \label{fig:all_distributions}
    \vspace{-2pt}
\end{figure*}
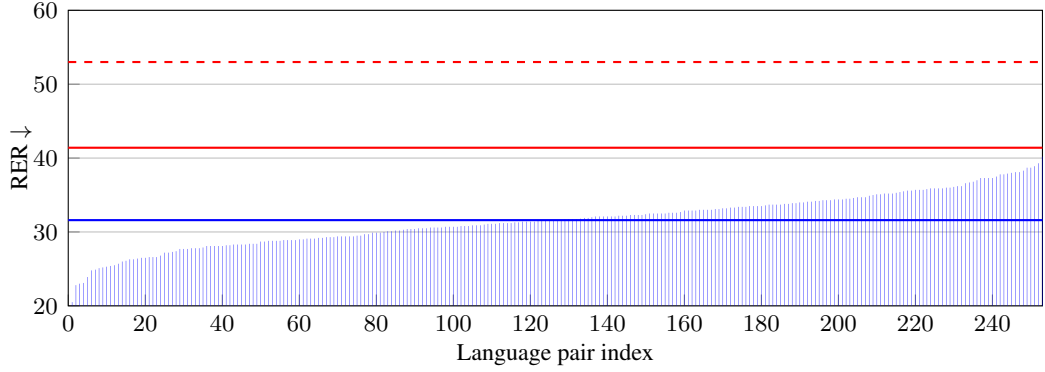
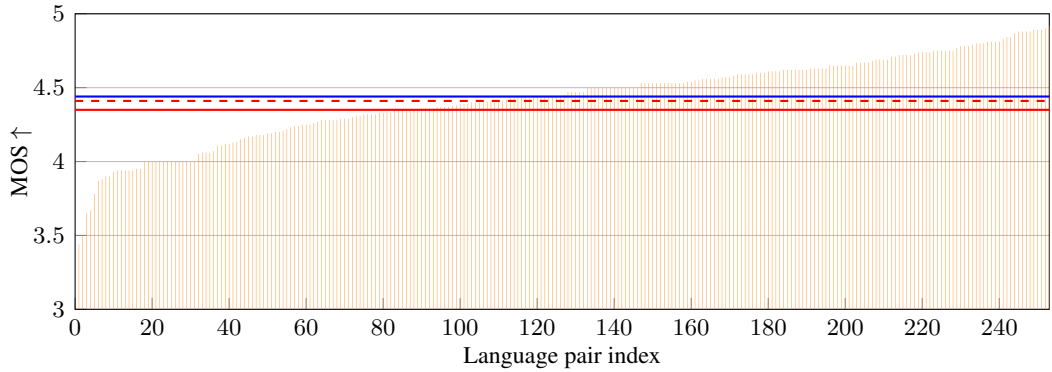
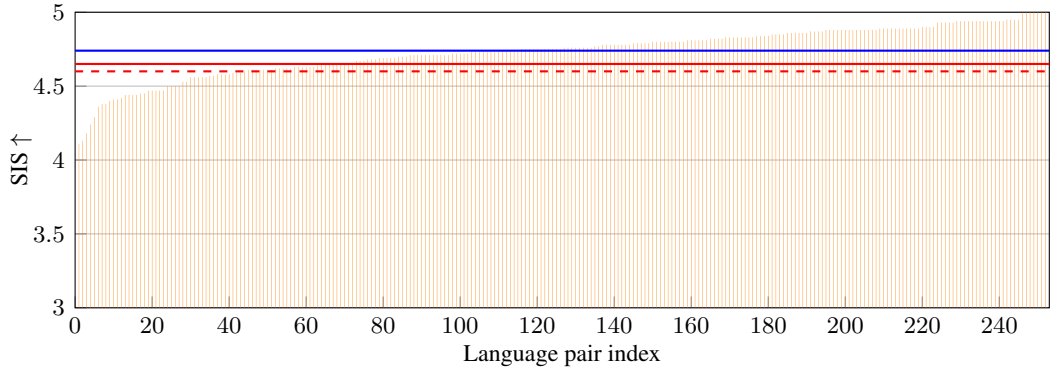

%% file: table/abl_vc.tex
\begin{table*}[]
\centering
% \small
\begin{tabular}{l|l|ccccc} \toprule
Model                        & Vocoder  & RER $\downarrow$ & NISQA $\uparrow$ & SECS $\uparrow$ & RTF $\downarrow$ & Relative Speed $\uparrow$ \\ \midrule\midrule
kNN-VC                       & HiFi-GAN &  25.0   &   \underline{4.35}    &   \underline{0.70}   &  \textbf{0.01}   &      \textbf{$\times$27.00}      \\ \midrule
\multirow{2}{*}{Diff-HierVC} & HiFi-GAN &  20.4   &    4.12   &   0.51   &  \underline{0.18}   &      \underline{$\times$1.50}      \\
                             & BigVGAN  &  \underline{19.7}   &   4.23    &   0.50   &  0.19   &      $\times$1.42      \\ \midrule
  \multirow{1}{*}{SeedVC}      & BigVGAN  &  \textbf{17.5}   &   \textbf{4.59}    &   \textbf{0.75}   &  0.27   &      $\times$1.00      \\ \bottomrule
\end{tabular}
\caption{Comparison of voice conversion models under cross-lingual settings.
NISQA scores are obtained using the NISQA-v2~\cite{nisqa} model, while speaker embedding cosine similarity (SECS) is computed based on embeddings from a pre-trained ECAPA-TDNN~\cite{ecapa} model trained on VoxCeleb~\cite{voxceleb}. For each metric, the best performance is indicated in bold, and the second-best is underlined.}
\label{tab_abl_vc}
\vspace{-2pt}
\end{table*}